\documentclass[runningheads]{llncs}

\usepackage{eccv}

\usepackage{eccvabbrv}

\usepackage{graphicx}
\usepackage{booktabs}
\usepackage{multirow} % 表格多行 \multirow
\usepackage{amssymb} % \mathbb
\usepackage{wrapfig} % wrapfig  浮动图不用占用一整行的

\usepackage{pifont} % \xmark \cmark
\newcommand{\xmark}{\color[rgb]{1,0,0}\ding{55}}
\newcommand{\cmark}{\color[rgb]{0,1,0}\ding{51}}

\usepackage[table,xcdraw]{xcolor} % \cellcolor
\definecolor{yellow}{rgb}{1, 1, 0.7}
\definecolor{orange}{rgb}{1, 0.85, 0.7}
\definecolor{red}{rgb}{1, 0.7, 0.7}

\usepackage[accsupp]{axessibility}  % Improves PDF readability for those with disabilities.

\usepackage[pagebackref,breaklinks,colorlinks,citecolor=eccvblue]{hyperref}
\usepackage{orcidlink}

\begin{document}

\title{StereoGS: Sparse-View 3D Gaussian Splatting via Stereo Priors} 

\titlerunning{StereoGS}

\author{Wenhao Yuan\orcidlink{0009-0005-5985-761X} \and
Yiyuan Ge\orcidlink{0009-0006-5442-1865} \and
Deli Cai\thanks{Corresponding author.}\orcidlink{0009-0000-0179-2781}}

\authorrunning{W. Yuan et al.}

\institute{South China University of Technology, China \\ \email{wenhaoyuan.stringer@gmail.com, yiyuange@mail.scut.edu.cn, eecaideli@mail.scut.edu.cn}}

\maketitle

\begin{abstract}
3D Gaussian Splatting (3DGS) has achieved remarkable success in real-time novel view synthesis, yet it suffers from severe overfitting under sparse-view settings due to insufficient geometric constraints. While recent methods introduce monocular depth priors to mitigate this, they inherently struggle with scale ambiguity and cross-view inconsistency, leading to defective geometry. In this paper, we propose StereoGS, a novel sparse-view 3DGS framework that integrates stereo priors to establish reliable binocular consistency. Unlike scale-agnostic monocular constraints, StereoGS introduces a Stereo Depth Regularization by constructing virtual stereo pairs during optimization and leveraging a foundation stereo model to enforce absolute scale and binocular-consistent structures. To further suppress overfitting and eliminate redundant primitives, we design a zero-overhead, plug-and-play Gradient-Aware Opacity Decay strategy that dynamically penalizes Gaussians based on their relative opacity gradient magnitudes. Combined with a Consistency-Aware Dense Initialization using zero-shot multi-view depth estimation, StereoGS effectively anchors primitives to accurate scene surfaces. Extensive experiments on LLFF, DTU, Mip-NeRF360, and Blender datasets demonstrate that StereoGS achieves state-of-the-art performance in sparse-view settings without incurring any additional inference overhead. Project Page: \url{https://stringerywh00.github.io/StereoGS_project_page/}

\end{abstract}

\section{Introduction} \label{sec:intro}

Novel View Synthesis (NVS) \cite{niemeyer2022regnerf, wang2023sparsenerf, barron2021mip} aims to synthesize realistic images from unseen viewpoints. While Neural Radiance Field (NeRF)~\cite{mildenhall2021nerf} and 3D Gaussian Splatting (3DGS)~\cite{kerbl20233dgs} have revolutionized high-fidelity NVS, 3DGS has gained particular attention for its real-time rendering capabilities. However, vanilla 3DGS relies heavily on dense input views. Under sparse-view settings \cite{niemeyer2022regnerf, wang2023sparsenerf, zhu2023FSGS, han2024binocular}, limited view overlap and insufficient appearance correspondences restrict sufficient geometric constraints, inevitably leading to severe overfitting. Consequently, achieving robust sparse-view 3DGS remains a critical yet challenging task for real-world applications.

% % CoherentGS, FSGS, DNGaussian, Depth-regularized optimization, SparseGS, bao2025loopsparsegs, song2025Depth_and_Density
To mitigate this, existing methods for sparse-view 3DGS \cite{chung2024Depth_regularized, li2024dngaussian, zhu2023FSGS} attempt to introduce geometric priors by using monocular depth estimation for geometry regularization. While effective, we argue that relying solely on monocular cues suffers from two fundamental limitations.
First, monocular depth inherently suffers from scale ambiguity, producing relative rather than absolute depth estimates. Since these relative depth maps lack a unified metric scale, enforcing them as regularization often conflicts with photometric consistency derived from the input images, thereby misguiding the optimization and leading to defective geometry.
Second, monocular depth maps lack cross-view consistency because they are inferred independently for each view. Consequently, the same spatial point is frequently assigned inconsistent depth values across different viewpoints. When back-projected into 3D space, these view-wise inconsistencies lead to spatial misalignments, destabilizing the Gaussian optimization and causing severe artifacts.
 
Recent studies \cite{han2024binocular, xu2024mvpgs, zheng2025nexusgs} attempt to integrate cross-view consistency for Gaussian optimization.
For instance, Binocular3DGS~\cite{han2024binocular} applies photometric loss on binocular images via the rendered depth map. However, this indirect pixel-level constraint lacks direct gradient guidance to 3D Gaussian primitives, failing to anchor them accurately to scene surfaces. NexusGS~\cite{zheng2025nexusgs} leverages the traditional epipolar geometry for initialization, but relying solely on initialization without continuous binocular consistency constraints during optimization still results in geometric degradation, especially in textureless or occluded regions. 

In this paper, we propose \textbf{StereoGS}, a sparse-view Gaussian Splatting framework that integrates stereo priors to establish reliable binocular consistency during Gaussian optimization. StereoGS is optimized via stereo priors during both initialization and optimization, compelling the 3D Gaussians to learn scale-accurate geometry. 
Specifically, unlike previous methods that rely on scale-agnostic monocular depth constraints, StereoGS regularizes the geometry by imposing a novel stereo depth constraint during optimization. For each training view, we construct a virtual right camera to form a stereo pair and render the right-view image, and depth maps for training views. By feeding the RGB pair of the ground-truth left-view and the rendered right-view into a foundation stereo model \cite{wen2025foundationstereo}, we obtain a highly accurate reference depth map that serves as an external stereo depth prior. We then perform supervision between the rendered and reference depth maps. This process effectively integrates a scale-aware stereo depth prior into the optimization, compelling the 3D Gaussians to learn absolute scale and binocular-consistent structures.

To suppress overfitting and eliminate redundant primitives, we introduce a general gradient-aware opacity decay strategy. Our core insight is that the opacity gradient of a Gaussian inherently reflects its necessity for scene reconstruction. Therefore, we dynamically penalize Gaussians based on their gradient magnitudes: primitives with smaller gradients are deemed less critical for rendering and receive a stronger decay penalty, while those with larger update demands are preserved. These mechanisms, alongside the stereo priors, serve as structural signals that backpropagate through the rendering pipeline to Gaussian primitives. As a result, StereoGS learns highly accurate, binocular-consistent geometric representations, directly mitigating the severe overfitting inherent in sparse-view settings.

Additionally, for Gaussian initialization, we employ a pre-trained multi-view depth estimator \cite{izquierdo2025mvsanywhere} combined with strict consistency filtering to estimate the multi-view-consistent depth maps for each view. By further applying cross-view reprojection errors, we eliminate outliers in the depth maps and fuse them into a dense point cloud, which serves as a robust initialization.

% 这一段可以根据排版删掉。
Notably, because the approaches in StereoGS are training-time strategies, employing StereoGS incurs no additional computational overhead during inference compared to the vanilla 3DGS.
Extensive experiments on LLFF, DTU, Mip-NeRF360, and Blender datasets demonstrate that StereoGS achieves state-of-the-art performance in sparse-view settings, effectively balancing geometric accuracy with rendering quality. 

Our main contributions are summarized as follows:
\begin{itemize}
    \item We introduce a Stereo Depth Regularization that utilizes stereo priors to enforce absolute scale constraints during optimization, fundamentally resolving the scale ambiguity inherent in monocular depth priors.
    \item We design Gradient-Aware Opacity Decay to dynamically adjust opacity during training based on the gradient, effectively eliminating redundant primitives and suppressing overfitting to ensure a compact and robust scene representation.
    \item Extensive experiments on LLFF, DTU, Mip-NeRF360, and Blender datasets demonstrate that our method achieves state-of-the-art results compared to existing sparse-view methods.
\end{itemize}

\section{Related Works}

\subsection{Radiance Fields}

Radiance Fields are employed for reconstructing 3D scenes and synthesizing
novel views. Neural Radiance Fields (NeRFs) \cite{mildenhall2021nerf} have seen significant advances that learn neural volumetric representations of 3D scenes and render images via volume rendering. While NeRF enables high-quality reconstruction from images, it requires high computational costs and suffers from slow training and inference speeds. Subsequent research has focused on addressing these bottlenecks by improving the rendering quality\cite{barron2021mip, barron2022mip360}, computational efficiency \cite{mueller2022instantngp, dvgo}, frequency regularization \cite{yang2022freenerf}.

A significant breakthrough in achieving real-time rendering is 3D Gaussian Splatting (3DGS) \cite{kerbl20233dgs}, which represents scenes using a set of explicit 3D Gaussian primitives. 3DGS reconstructs high-quality scenes rapidly by utilizing differentiable splatting, excelling particularly in handling high-frequency details and offering intuitive interpretability. Building on its significant performance, many follow-up works focus on improving the rendering details \cite{ye2024absgs, zhang2024pixel, kheradmand20243d, yan2024multi}, compressing the Gaussians and accelerating rendering \cite{lee2024compact, niedermayr2024compressed}.

\subsection{Novel View Synthesis from Sparse View}

NeRF-based and 3DGS-based methods have demonstrated impressive rendering performance, whereas in real-world scenarios, only a small number of images are available (i.e., sparse input views). Directly applying the methods in a dense-view setting leads to performance degradation due to overfitting.
Early studies have investigated the regularization of NeRFs under sparse-view conditions \cite{yang2022freenerf, niemeyer2022regnerf, kim2022infonerf, deng2022dsnerf}. Other approaches \cite{zhou2023sparsefusion, wynn2023diffusionerf} leverage generative models that are pre-trained on large-scale datasets to distill generative prior into 3D representation. In contrast, alternative strategies \cite{deng2022dsnerf,roessle2022dense,song2023darf,wang2023sparsenerf} leverage external geometric priors derived from pre-trained models (e.g., depth priors) to guide and regularize the training process.

%%%%%%%%%%%% 3DGS的
Recently, 3D Gaussian Splatting has been adapted to sparse settings, with a primary focus on initialization, depth regularization, and structural constraints.  
% 单目深度初始化
For Gaussian initialization, Chung \textit{et al.} \cite{chung2024Depth_regularized} and CoherentGS \cite{paliwal2024coherentgs} estimate monocular depth maps and refine them to achieve more accurate initialization. 
% 单目深度正则化 DNGaussian, FSGS
DNGaussian \cite{li2024dngaussian} and FSGS \cite{zhu2023FSGS} apply depth regularization based on monocular depth estimators.
However, relying on monocular depth often lacks absolute scale depth perception.
% MVS深度初始化
Although MVPGS \cite{xu2024mvpgs} leverages Multi-View Stereo (MVS) depth estimation to obtain a well-initialized, view-consistent point cloud, the absence of continuous geometric regularization during optimization causes this high-quality initial structure to gradually deteriorate.
% 其他方法，CoR-GS和DrouGaussian
Other approaches, such as Gaussian pruning based on point disagreement \cite{zhang2024cor-gs} and a random dropout strategy \cite{park2025dropgaussian}, also effectively mitigate overfitting to known yet sparse views.
While several methods have explored integrating stereo matching models for dense-view settings \cite{safadoust2024self} or surface reconstruction \cite{zhang2026eve3d, gu2026sparsesurf}, their potential in sparse-view novel view synthesis remains largely underexplored.
% 涉及到双目的
To integrate cross-view consistency during optimization, Binocular3DGS \cite{han2024binocular} warps rendered RGB images and applies a photometric loss between the constructed binocular RGB images. We argue that this self-supervision constraint on RGB images lacks explicit geometric constraints and leads to geometric artifacts.
Despite improvements, existing methods predominantly lack explicit cross-view consistency constraints during Gaussian optimization and can lead to inevitable geometric artifacts. In contrast, we propose integrating a stereo constraint from the stereo matching prior into the optimization process.

\section{Method}

\subsection{Preliminary for 3D Gaussian Splatting}
\label{sec:preliminary}

3D Gaussian Splatting (3DGS)~\cite{kerbl20233dgs} models the 3D scene as a collection of anisotropic 3D Gaussians, serving as an explicit scene representation. Each primitive $G_i$ is spatially defined by a center position $\mu$ and a 3D covariance matrix $\Sigma$. The influence of a Gaussian at a spatial coordinate $x$ is formulated as:
\begin{equation}
    G(x) = \exp\left(-\frac{1}{2} (x - \mu)^\top \Sigma^{-1} (x - \mu)\right).
\end{equation}
To ensure the positive semi-definite property of the covariance matrix during optimization, $\Sigma$ is factorized into a scaling matrix $S$ and a rotation matrix $R$, such that $\Sigma = R S S^\top R^\top$. Additionally, each Gaussian encodes view-dependent appearance via spherical harmonic (SH) coefficients and a learnable opacity $\alpha$.

To synthesize novel views, these 3D Gaussians are projected into the 2D image plane via differentiable splatting. Given a viewing transformation $W$ and the Jacobian of the affine approximation of the projective transformation $J$, the 3D covariance $\Sigma$ is transformed into a 2D planar covariance $\Sigma' = J W \Sigma W^\top J^\top$. The rendering pipeline employs a tile-based rasterizer that sorts the Gaussians overlapping a specific pixel by depth. The final pixel color $C$ is derived through $\alpha$-blending, accumulating the contribution of $N$ ordered Gaussians:
\begin{equation}
    C = \sum_{i=1}^{N} c_i \alpha'_i \prod_{j=1}^{i-1} (1 - \alpha'_j),
\end{equation}
where $c_i$ denotes the color decoded from SH coefficients, and $\alpha'_i$ represents the effective 2D opacity. 

Similarly, the final pixel depth $D$ can be rendered by accumulating the projected depth values $d_i$ (the distance from the camera center to the Gaussian center) using the same blending weights:
\begin{equation}
    D = \sum_{i=1}^{N} d_i \alpha'_i \prod_{j=1}^{i-1} (1 - \alpha'_j).
\end{equation}

3D Gaussian Splatting is typically trained end-to-end by minimizing the discrepancy between the rendered and ground-truth images using a combination of $\mathcal{L}_1$ and D-SSIM loss terms.

\begin{figure}[t] 
    \vspace{0mm}
    \centering
       \includegraphics[width=1.0\linewidth]{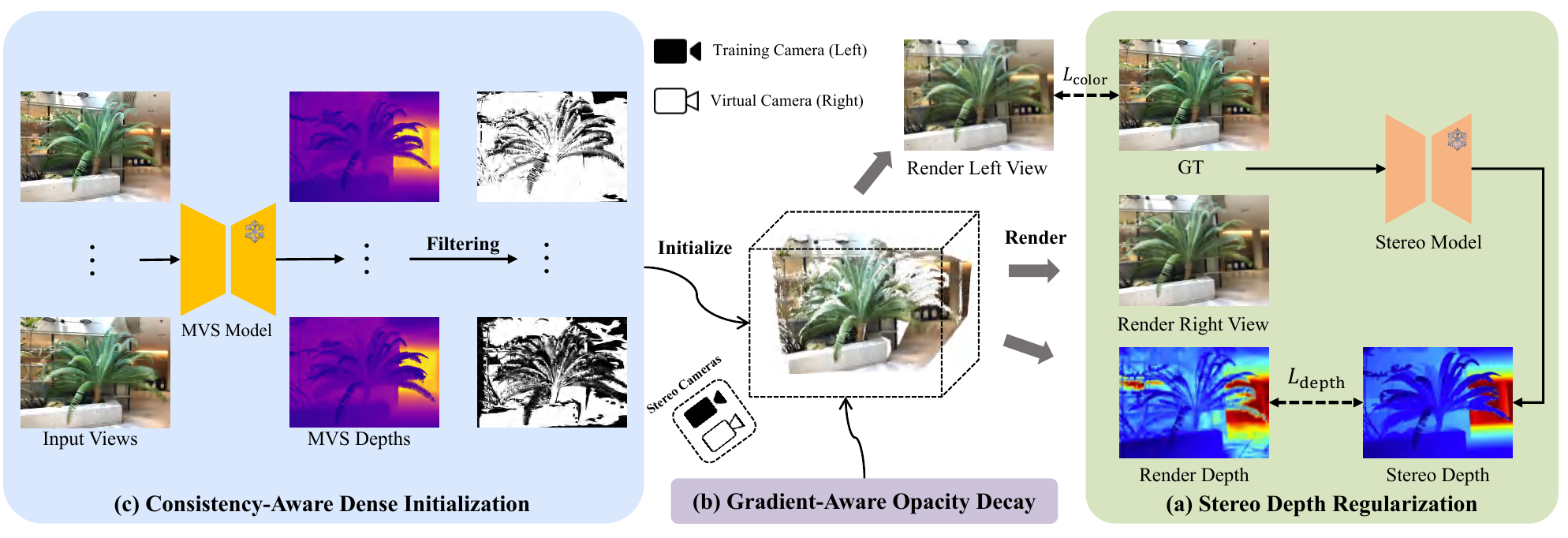}
       \vspace{-2mm}
       \caption{\textbf{Overview of StereoGS}, comprising three components:  (a) Stereo Depth Regularization synthesizes virtual stereo pairs to extract stereo depth, enforcing explicit geometric consistency between paired views during optimization. (b) Gradient-Aware Opacity Decay dynamically penalizes Gaussian opacities according to their gradients, facilitating the removal of redundant primitives and mitigating overfitting. (c) Consistency-Aware Dense Initialization leverages multi-view depth priors and geometric filtering to construct a dense and reliable geometric foundation.}
    \label{fig:overview}
\end{figure}

%%%%%%%%%%%%%%%%%%%%%%%%%%%%%%%%%%%%%
%%%%%%%%%%%%%%%%%%%%%%%%%%%%%%%%%%%%%
%%%%%%%%%%%%%%%%%%%%%%%%%%%%%%%%%%%%%

\subsection{Stereo Depth Regularization}
\label{sec:stereo_reg}
Geometric inaccuracy remains a critical bottleneck in sparse-view 3D Gaussian Splatting. Several prior works \cite{li2024dngaussian, zhu2023FSGS} incorporate monocular depth~\cite{ranftl2020towards, ranftl2021vision, yang2024depth} as a regularization term during the optimization process. However, the reconstructed geometry often remains suboptimal due to the inherent scale ambiguity and cross-view inconsistency of these monocular priors.
To address this limitation, recent approaches~\cite{han2024binocular} enforce stereo consistency using photometric constraints. They warp synthesized views to the input views using the rendered depth to minimize reconstruction errors. 
However, this warping-based approach lacks direct supervision on Gaussian geometry and is prone to failure in textureless or repetitive regions.

To overcome these limitations, we propose stereo depth regularization. By constructing virtual stereo pairs and leveraging a foundation stereo model~\cite{wen2025foundationstereo}, we derive reliable stereo depth to regularize the Gaussian geometry, as illustrated in Fig.~\ref{fig:overview} (a).

\noindent\textbf{Stereo Camera Pair Construction.}
During optimization, we treat the training view as the left camera and synthesize a virtual right camera by applying a horizontal translation. We then render the corresponding right-view image $\hat{I}_{r}$ using the current 3D Gaussians, forming a stereo pair $(I_l, \hat{I}_r)$ with the ground-truth left image $I_l$.

\noindent\textbf{Left-Right Consistency Check.}
We employ FoundationStereo~\cite{wen2025foundationstereo} as our pre-trained stereo matching model $\Phi$. The left-view disparity is computed as $\hat{D}_l = \Phi(I_l, \hat{I}_r)$. We explicitly use the clean ground-truth left image instead of a rendered one because the estimated disparity serves as an absolute depth reference; avoiding optimization noise and rendering artifacts ensures a highly reliable geometric prior.

To filter out unreliable predictions, we apply the left-right consistency check that is widely applied in stereo vision literature~\cite{godard2017monodepth, 2008SGM, jie2018left_recurrent}. We simultaneously extract the right-view disparity. Since standard stereo models typically output left-view disparity, we introduce a horizontal flipping operator $\mathcal{F}(\cdot)$ to compute it: $\hat{D}_r = \mathcal{F}\left( \Phi \left( \mathcal{F}(I_l), \mathcal{F}(\hat{I}_r) \right) \right)$.
The left-right consistency effectively identifies occluded regions and eliminates false matches by judging whether the disparity discrepancy between $\hat{D}_l$ and $\hat{D}_r$ exceeds a threshold, forming an occlusion mask $M_{\text{occ}}$. For specific details, please refer to the supplementary material.

To further refine the valid regions, we compute a background mask $M_\text{bg}$ and a disparity anomaly mask $M_\text{anomaly}$. The final validity mask for depth regularization is derived by fusing these masks: $M_{\text{valid}} = 1 - (M_{\text{bg}} \lor M_{\text{occ}} \lor M_{\text{anomaly}})$.

\noindent\textbf{Depth Regularization.}
We convert the estimated disparity $\hat{D}_l$ into depth $Z_\text{stereo}$ using the standard triangulation formula: $Z_\text{stereo} = {f d}/{\hat{D}_l}$, where $f$ is the focal length and $d$ is the baseline between camera pairs. This $Z_{\text{stereo}}$ serves as pseudo-ground-truth and the stereo prior from the external stereo model.
To further improve numerical stability during loss computation and to better constrain the geometric accuracy in near-foreground regions, we minimize the $L_1$ error in the inverse depth space between $Z_\text{stereo}$ and the rendered left-view depth $\hat{Z}$ for all valid pixels, using the validity mask $M_{\text{valid}}$. 

\begin{equation}
    \mathcal{L}_{\text{depth}} = \lVert M_{\text{valid}} \odot (\frac{1}{\hat{Z}} - \frac{1}{Z_\text{stereo}}) \rVert_1,
    \label{eq:depth_loss}
\end{equation}

\subsection{Gradient-Aware Opacity Decay}
\label{sec:adaptive_decay}

Opacity determines the visibility and contribution of each Gaussian primitive to the rendered image. Despite the importance of this attribute, existing methods~\cite{kerbl20233dgs, chung2024Depth_regularized, li2024dngaussian, zhu2023FSGS, xu2024mvpgs, zheng2025nexusgs} predominantly rely on a periodic opacity reset mechanism to eliminate noisy primitives. While effective in dense-view scenarios, this aggressive reset indiscriminately suppresses both stable surface structures and transient floaters in sparse-view settings. Although recent sparse-view techniques~\cite{han2024binocular} employ a strategy with fixed decay rate that penalizes all Gaussians equally, this approach cannot effectively evaluate the importance of individual Gaussians or distinguish useful geometry from noise.

Motivated by these limitations, we propose a gradient-aware opacity decay strategy. Our core insight is that the opacity gradient magnitude indicates a Gaussian primitive's importance: Gaussians with large opacity gradients contribute significantly to reducing photometric error and should be preserved, whereas those with negligible gradients likely represent redundant floaters or noise.

To implement this insight, we derive a dynamic decay factor $\gamma$ by incorporating the local gradient magnitude into a predefined base factor $\gamma_\text{base}$. This factor $\gamma$ is then used to modulate the original opacity $\alpha$. 
Formally, let $g = |\nabla_{\alpha} \mathcal{L}|$ denote the magnitude of the gradient of the loss function $\mathcal{L}$ with respect to the opacity $\alpha$. We observe that the absolute values of $g$ are typically minuscule (e.g., on the order of $10^{-6}$). To ensure robustness against such scale variations, we guide the opacity decay using a relative gradient magnitude. Inspired by Group Relative Policy Optimization (GRPO)~\cite{shao2024deepseekmath}, which evaluates the output actions based on their relative advantages within a group rather than relying on an absolute value network, we compute a relative gradient $\beta = g / \bar{g}$ for each Gaussian, where $\bar{g}$ is the mean opacity gradient magnitude across all Gaussians at the current iteration. Incorporating a basic decay factor $\gamma_\text{base}$, we derive a dynamic decay factor $\gamma$ via an exponential soft-thresholding function: 
\begin{equation}
    \gamma = 1 - (1 - \gamma_\text{base}) \exp\left(-s \cdot \beta \right),
\end{equation}
where $s$ is a sensitivity hyperparameter controlling the decay curvature. Finally, the opacity is scaled by $\hat{\alpha} = \gamma \alpha$.

\subsection{Consistency-Aware Dense Initialization}
\label{sub_sec:gaussian_init}

Previous methods~\cite{kerbl20233dgs, chung2024Depth_regularized} typically utilize sparse point clouds generated by Structure-from-Motion (SfM)~\cite{schonberger2016structure} for initialization. However, in sparse-view settings, SfM often yields point clouds that are too sparse and noisy to adequately represent the scene. 
While some recent works~\cite{xu2024mvpgs, han2024binocular} have explored using MVS priors (e.g., MVS models~\cite{cao2022mvsformer} and keypoint matching networks~\cite{truong2023pdc}) to improve initialization, their performance is often constrained by the domain gap between pre-training data and diverse in-the-wild scenes, as illustrated in Fig. \ref{fig:overview} (c).

Building on this MVS-based paradigm, we overcome the generalization bottleneck by leveraging a more advanced zero-shot model, MVSAnywhere~\cite{izquierdo2025mvsanywhere}, to extract multi-view-consistent depth maps. Specifically, for each training view (treated as the target image), we input it along with the remaining source images into the MVS model to estimate its depth map. By iterating over all training views, we obtain a set of cross-view consistent depth maps.
We then adopt a geometric filtering strategy~\cite{yao2018mvsnet} based on cross-view reprojection errors to eliminate outliers. The filtered depth maps are finally back-projected and fused into a unified dense point cloud. Thanks to the robust zero-shot capabilities of MVSAnywhere, we obtain higher-quality point clouds compared to~\cite{xu2024mvpgs} and~\cite{han2024binocular}, providing a significantly more robust initialization for the 3D Gaussians. We provide visual comparisons of the initialization point clouds in the experiment section.

\subsection{Training Loss}
The total objective function comprises two components: the proposed stereo matching prior-guided depth regularization $\mathcal{L}_{\text{depth}}$, as detailed in Eq.~(\ref{eq:depth_loss}), and the color reconstruction loss $\mathcal{L}_{\text{color}}$ adopted from 3DGS~\cite{kerbl20233dgs}. We define the overall loss $\mathcal{L}$ as:
\begin{equation}
    \mathcal{L} = \mathcal{L}_{\text{color}} + \mathcal{L}_{\text{depth}},
\end{equation}
where $\mathcal{L}_{\text{color}}$ represents a linear combination of the $\mathcal{L}_1$ loss and the D-SSIM loss $\mathcal{L}_{\text{D-SSIM}}$, formulated as:
\begin{equation}
    \mathcal{L}_{\text{color}} = (1 - \lambda) \mathcal{L}_1 + \lambda \mathcal{L}_{\text{D-SSIM}}.
\end{equation}

\section{Experiments}

\subsection{Experimental Settings}

\noindent\textbf{Datasets.} We conduct experiments on four public datasets: LLFF \cite{mildenhall2019llff}, DTU \cite{jensen2014dtu}, Mip-NeRF360~\cite{barron2022mip360}, and Blender~\cite{mildenhall2021nerf}. Following prior works \cite{han2024binocular, li2024dngaussian, zhu2023FSGS, niemeyer2022regnerf}, we used 3, 6, and 9 views as training sets for the LLFF and DTU datasets, 12 and 24 views for the Mip-NeRF360 dataset, and 8 images for training on the Blender dataset. The selection of test images remained consistent with previous works \cite{han2024binocular, li2024dngaussian, zhu2023FSGS, niemeyer2022regnerf}. The downsampling rates for the LLFF, DTU, Mip-NeRF360, and Blender datasets are 8, 4, 8, and 2, respectively.

\noindent\textbf{Baselines.} We compare the proposed StereoGS with representative existing methods, including NeRF-based approaches (RegNeRF \cite{niemeyer2022regnerf}, FreeNeRF \cite{yang2022freenerf}, and SparseNeRF \cite{wang2023sparsenerf}) and 3DGS-based approaches (3DGS \cite{kerbl20233dgs}, DNGaussian \cite{li2024dngaussian}, FSGS \cite{zhu2023FSGS}, CoR-GS \cite{zhang2024cor-gs}, MVPGS~\cite{xu2024mvpgs}, NexusGS~\cite{zheng2025nexusgs}, DropGaussian~\cite{park2025dropgaussian}, Binocular3DGS~\cite{han2024binocular}, and D$^2$GS~\cite{song2025Depth_and_Density}). For most evaluated methods, we directly report the best quantitative results from the respective original publications. For the vanilla 3DGS, we report the results obtained from our own implementation. We employ the average values of PSNR, SSIM~\cite{wang2004image}, and LPIPS~\cite{zhang2018unreasonable} as the quantitative evaluation metrics.

\begin{table}[t]
    \scriptsize
    \caption{
        \textbf{Quantitative comparison on LLFF with 3, 6, 9 training views.} 
        We mark the \colorbox[RGB]{255,179,179}{best}, \colorbox[RGB]{255,217,179}{second best}, and \colorbox[RGB]{255,255,179}{third best} methods in cells, respectively. ``*'' denotes using random dropout proposed by \cite{park2025dropgaussian}.
    }
    \label{tab:llff}
    \resizebox{\linewidth}{!}{
        \begin{tabular}{l c c c c c c c c c}
        \toprule
        \multirow{2}{*}[-0.5ex]{Methods}
        & \multicolumn{3}{c}{3-view}
        & \multicolumn{3}{c}{6-view}
        & \multicolumn{3}{c}{9-view} \\ 
        
        & \multicolumn{1}{c}{PSNR$\uparrow$}
        & \multicolumn{1}{c}{SSIM$\uparrow$}
        & \multicolumn{1}{c}{LPIPS$\downarrow$}
        
        & \multicolumn{1}{c}{PSNR$\uparrow$}
        & \multicolumn{1}{c}{SSIM$\uparrow$}
        & \multicolumn{1}{c}{LPIPS$\downarrow$}
        
        & \multicolumn{1}{c}{PSNR$\uparrow$}
        & \multicolumn{1}{c}{SSIM$\uparrow$}
        & \multicolumn{1}{c}{LPIPS$\downarrow$} \\ 
        
        \midrule
        RegNeRF~\cite{niemeyer2022regnerf} & 19.08 & 0.587 & 0.336 & 23.10 & 0.760 & 0.206 & 24.86 &  0.820 & 0.161 \\
        FreeNeRF~\cite{yang2022freenerf} & 19.63 & 0.612 & 0.308 & 23.73 &  0.779 & 0.195  & 25.13 &  0.827 & 0.160 \\
        SparseNeRF~\cite{wang2023sparsenerf} & 19.86 & 0.624 & 0.328 & 23.26 & 0.741 & 0.235 & 24.27 & 0.781 & 0.228 \\
        
        \midrule
        3DGS~\cite{kerbl20233dgs} & 16.02 & 0.465 & 0.378 & 19.45 & 0.627 & 0.268 & 21.13 & 0.715 & 0.214 \\
        DNGaussian~\cite{li2024dngaussian} & 19.12 &  0.591 &  0.294 & 22.18 & 0.755 &  0.198 & 23.17 & 0.788 & 0.180 \\
        FSGS~\cite{zhu2023FSGS} & 20.31 & 0.652 & 0.288 & 24.20 & 0.811 & 0.173 & 25.32 & 0.856 & 0.136 \\
        CoR-GS~\cite{zhang2024cor-gs} & 20.45 & 0.712 & 0.196 & 24.49 & 0.837 & 0.115 & 26.06 & 0.874 & 0.089 \\
        MVPGS \cite{xu2024mvpgs} & 20.54 & 0.727 & 0.194 & 23.64 & 0.819 & 0.137 & 24.23 & 0.843 & 0.120  \\
        DropGaussian \cite{park2025dropgaussian} & 20.76 & 0.713 & 0.200 & 24.74 & 0.837 & 0.117 & \cellcolor{yellow}26.21 & 0.874 & \cellcolor{yellow}0.088  \\
        NexusGS \cite{zheng2025nexusgs} & 21.07 & 0.738 & 0.177 & - & - & - & - & - & -  \\
        Binocular3DGS \cite{han2024binocular} & \cellcolor{yellow}21.44 & \cellcolor{yellow}0.751 & \cellcolor{yellow}0.168 & \cellcolor{yellow}24.87 & \cellcolor{yellow}0.845 & \cellcolor{yellow}0.106 & 26.17   & \cellcolor{yellow}0.877 & 0.090 \\
        D$^2$GS \cite{song2025Depth_and_Density} & 21.35 & 0.746 & 0.179 & 24.84 & 0.834 & 0.122 & - & - & -  \\
        Ours & \cellcolor{orange}21.91 & \cellcolor{orange}0.773 & \cellcolor{orange}0.157 & \cellcolor{orange}24.92 & \cellcolor{orange}0.853 & \cellcolor{orange}0.104 & \cellcolor{orange}26.25 & \cellcolor{orange}0.879 & \cellcolor{orange}0.087  \\
        Ours* & \cellcolor{red}22.05 & \cellcolor{red}0.783 & \cellcolor{red}0.147 & \cellcolor{red}25.40 & \cellcolor{red}0.856 & \cellcolor{red}0.102 & \cellcolor{red}26.44 & \cellcolor{red}0.883 & \cellcolor{red}0.082  \\
        \bottomrule
        \end{tabular}
    }
\end{table}

\begin{figure}[t!] 
    \vspace{0mm}
    \centering
       \includegraphics[width=\linewidth]{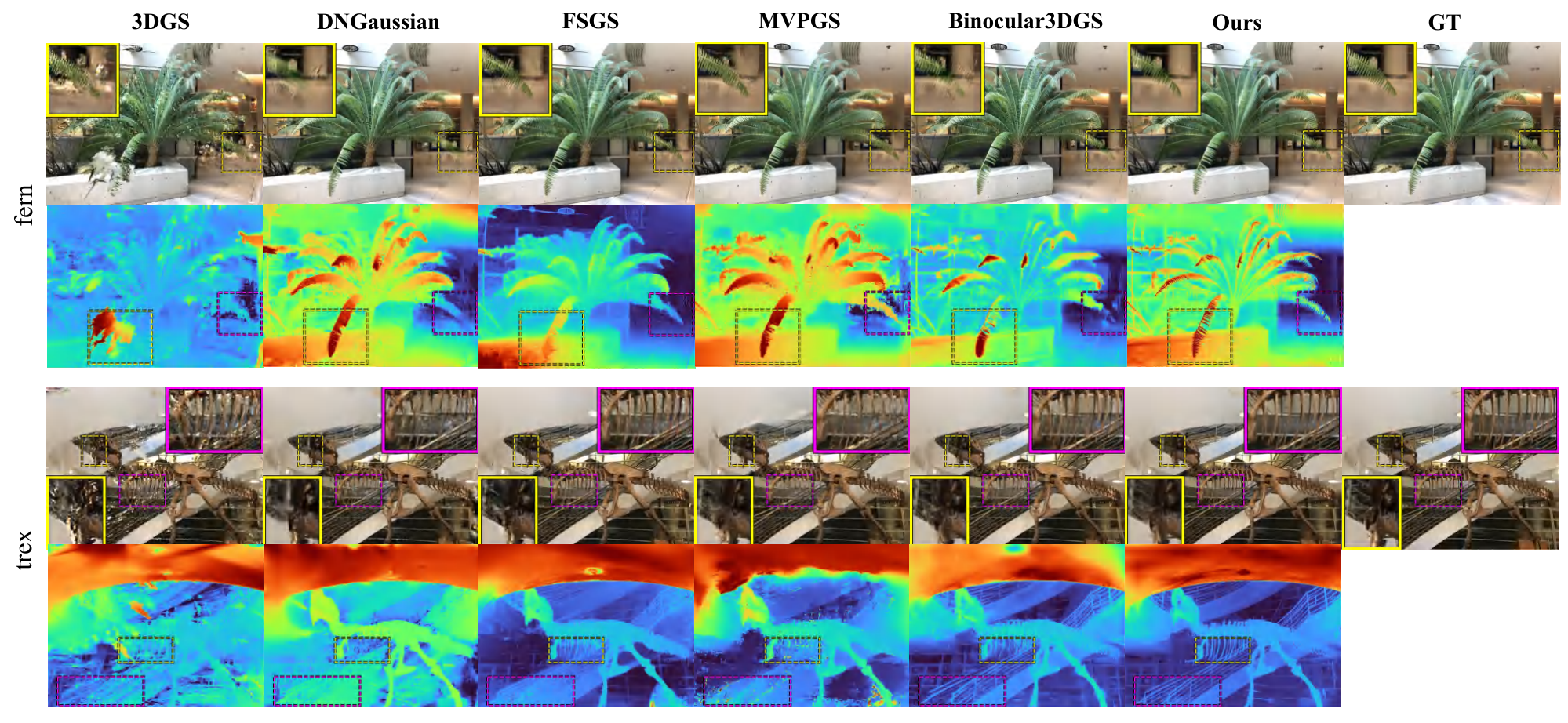}
       \caption{\textbf{Visual comparison on LLFF dataset, including RGB images and colored depth maps.} The Gaussians are learned under 1/8 resolution with 3 input views.}
    \label{fig:visual_compare_LLFF}
\end{figure}

\subsection{Comparison}

\begin{table}[t!]
    \scriptsize
    \caption{
        \textbf{Quantitative comparison on DTU with 3, 6, 9 training views.} 
         We mark the \colorbox[RGB]{255,179,179}{best}, \colorbox[RGB]{255,217,179}{second best}, and \colorbox[RGB]{255,255,179}{third best} methods in cells, respectively. ``*'' denotes using random dropout proposed by \cite{park2025dropgaussian}.
    }
    \label{tab:dtu}
    \resizebox{\linewidth}{!}{
        \begin{tabular}{l c c c c c c c c c}
        \toprule
        \multirow{2}{*}[-0.5ex]{Methods}
        & \multicolumn{3}{c}{3-view}
        & \multicolumn{3}{c}{6-view}
        & \multicolumn{3}{c}{9-view} \\ 
        
        & \multicolumn{1}{c}{PSNR$\uparrow$}
        & \multicolumn{1}{c}{SSIM$\uparrow$}
        & \multicolumn{1}{c}{LPIPS$\downarrow$}
        
        & \multicolumn{1}{c}{PSNR$\uparrow$}
        & \multicolumn{1}{c}{SSIM$\uparrow$}
        & \multicolumn{1}{c}{LPIPS$\downarrow$}

        & \multicolumn{1}{c}{PSNR$\uparrow$}
        & \multicolumn{1}{c}{SSIM$\uparrow$}
        & \multicolumn{1}{c}{LPIPS$\downarrow$} \\ 
        
        \midrule
        RegNeRF~\cite{niemeyer2022regnerf} & 18.89 & 0.745 & 0.190 & 22.20 & 0.841 & 0.117 & 24.93 & 0.884 & 0.089  \\
        FreeNeRF~\cite{yang2022freenerf} & 19.52 & 0.787 & 0.173 & 23.25 & 0.844 & 0.131 & 25.38 & 0.888 & 0.102 \\
        SparseNeRF~\cite{wang2023sparsenerf} & 19.47 & 0.829 & 0.183 & 23.26 & 0.843 & 0.135 & 25.13 & 0.871 & 0.114 \\
        
        \midrule
        3DGS~\cite{kerbl20233dgs} & 10.99 & 0.585 & 0.313 & 20.33 & 0.776 & 0.223 & 22.90 & 0.816 & 0.173 \\
        DNGaussian~\cite{li2024dngaussian} & 18.91 & 0.790 & 0.176 & 22.10 & 0.851 & 0.148 & 23.94 & 0.887 & 0.131  \\
        FSGS~\cite{zhu2023FSGS} & 17.34 & 0.818 & 0.169 & 21.55 & 0.880 & 0.127 & 24.33 & 0.911 & 0.106 \\
        CoR-GS~\cite{zhang2024cor-gs} & 19.21 & 0.853 & 0.119 & \cellcolor{yellow}24.51 & 0.917 & 0.068 & \cellcolor{orange}27.18 & \cellcolor{orange}0.947 & \cellcolor{red}0.045 \\
        MVPGS \cite{xu2024mvpgs} & 20.65 & \cellcolor{yellow}0.877 & \cellcolor{orange}0.099 & 23.98 & \cellcolor{yellow}0.921 & \cellcolor{yellow}0.066 & 26.45 &  \cellcolor{yellow}0.946 & \cellcolor{orange}0.047  \\
        NexusGS \cite{zheng2025nexusgs} & 20.21 & 0.869 & \cellcolor{yellow}0.102 & - & - & - & - & - & -  \\
        Binocular3DGS \cite{han2024binocular} & \cellcolor{yellow}20.71 & 0.862 & 0.111 & 24.31 & 0.917 & 0.073 & 26.70 & \cellcolor{orange}0.947 & 0.052  \\
        Ours & \cellcolor{orange}21.46 & \cellcolor{orange}0.879 & \cellcolor{orange}0.099 & \cellcolor{orange}24.86 & \cellcolor{orange}0.926 & \cellcolor{orange}0.064 & \cellcolor{yellow}26.83 & \cellcolor{red}0.955 & \cellcolor{yellow}0.049  \\
        Ours* & \cellcolor{red}22.00 & \cellcolor{red}0.890 & \cellcolor{red}0.095 & \cellcolor{red}25.41 & \cellcolor{red}0.938 & \cellcolor{red}0.063 & \cellcolor{red}27.39 & \cellcolor{red}0.955 & \cellcolor{orange}0.047  \\
        \bottomrule
        \end{tabular}
    }
\end{table}

\begin{figure}[t!] 
    \vspace{0mm}
    \centering
       \includegraphics[width=1.0\linewidth]{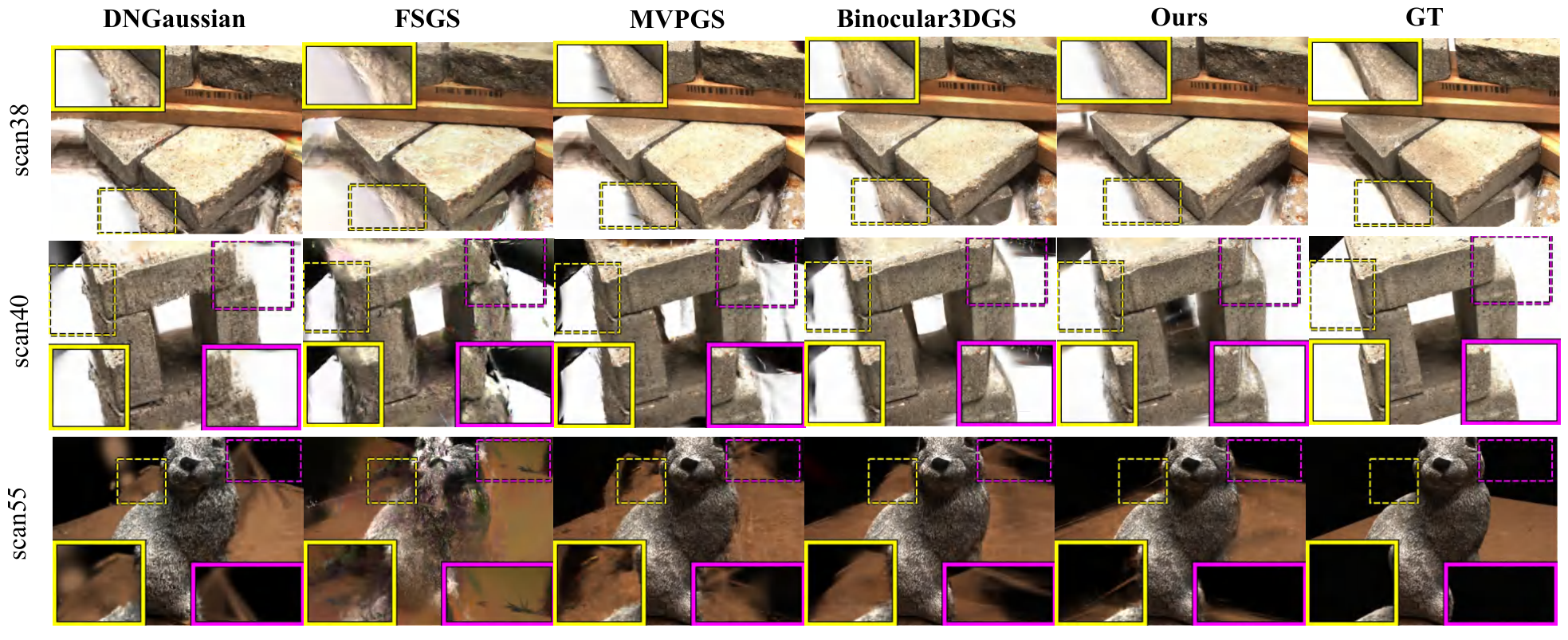}
       \caption{\textbf{Visual comparison on DTU under 1/8 resolution with 3 views.}}
    \label{fig:visual_compare_DTU}
\end{figure}

\noindent\textbf{Comparison on LLFF.} 
Table \ref{tab:llff} presents the quantitative results on the LLFF dataset under configurations of 3, 6, and 9 input views. Our StereoGS consistently outperforms prior approaches across all view configurations in terms of PSNR, SSIM, and LPIPS. For NexusGS, we could not reproduce their results and report only the 3-view result from their paper. Fig. \ref{fig:visual_compare_LLFF} provides visual comparisons of novel view synthesis and depth rendering for the \textit{fern} and \textit{trex} scenes from the LLFF dataset. As demonstrated by the rendered RGB images, vanilla 3DGS~\cite{kerbl20233dgs} captures only the coarse scene layout and suffers from severe patchy artifacts. For methods that employ monocular depth regularization, such as DNGaussian~\cite{li2024dngaussian} and FSGS~\cite{zhu2023FSGS}, the primary limitation lies in an inability to accurately reconstruct the underlying scene geometry. Although MVPGS~\cite{xu2024mvpgs} and Binocular3DGS~\cite{han2024binocular} utilize MVS to establish an improved geometric initialization, these approaches still struggle during the optimization process. Specifically, Binocular3DGS relies on a self-supervised photometric consistency loss that is inherently unreliable in textureless regions. Furthermore, the use of a fixed-ratio opacity decay rate in Binocular3DGS fails to preserve essential surface Gaussians, as shown by the skeleton of the \textit{trex}. These limitations result in geometric distortion and subsequent rendering artifacts, ultimately degrading the reconstructed details. In contrast, our StereoGS utilizes stereo depth regularization and employs a gradient-aware opacity decay strategy, effectively resolving ambiguities in textureless regions and preserving critical scene geometry.

\begin{table}[t]
  \caption{
        \textbf{Quantitative results on Mip-NeRF360 with 12 and 24 views.}
    }
  \label{tab:mipnerf360} 
  \centering
  \renewcommand{\arraystretch}{0.9} % 调整行间距
  \resizebox{0.85\linewidth}{!}{
      \setlength{\tabcolsep}{2mm}
      \begin{tabular}{l c c c c c c} 
        \toprule
        \multirow{2}{*}[-0.5ex]{Methods}
        & \multicolumn{3}{c}{12-view}
        & \multicolumn{3}{c}{24-view} \\ 

        & \multicolumn{1}{c}{PSNR$\uparrow$}
        & \multicolumn{1}{c}{SSIM$\uparrow$}
        & \multicolumn{1}{c}{LPIPS$\downarrow$}
        
        & \multicolumn{1}{c}{PSNR$\uparrow$}
        & \multicolumn{1}{c}{SSIM$\uparrow$}
        & \multicolumn{1}{c}{LPIPS$\downarrow$} \\ 
    
        \midrule
        RegNeRF~\cite{niemeyer2022regnerf} & 18.55 & 0.524 & 0.426 & 22.19 & 0.643 & 0.335 \\
        FreeNeRF~\cite{yang2022freenerf} & 18.68 & 0.528 & 0.421 & 22.78 & 0.689 & 0.323 \\
        SparseNeRF~\cite{wang2023sparsenerf} & 18.73 & 0.531 & 0.419 & 22.85 & 0.693 & 0.315 \\
        
        \midrule
        3DGS~\cite{kerbl20233dgs} & 18.52 & 0.523 & 0.415 & 22.80 & 0.708 & 0.276 \\
        FSGS~\cite{zhu2023FSGS} & 18.80 & 0.531 & 0.418 & 23.28 & 0.715 & 0.274 \\
        CoR-GS~\cite{zhang2024cor-gs} & 19.52 & 0.558 & 0.418 & 23.39 & 0.727 & 0.271 \\
        NexusGS~\cite{zheng2025nexusgs} & - & - & - & 23.86 & 0.753 & \cellcolor{red}0.206 \\ 
        DropGaussian~\cite{park2025dropgaussian} & 19.74 & 0.577 & 0.364 & 24.05 & 0.761 & 0.226 \\ 
        D$^2$GS \cite{song2025Depth_and_Density} & \cellcolor{yellow}20.09 & \cellcolor{yellow}0.587 & \cellcolor{yellow}0.356 & \cellcolor{yellow}24.13 & \cellcolor{yellow}0.763 & \cellcolor{orange}0.221  \\
        Ours & \cellcolor{orange}20.25 & \cellcolor{orange}0.618 & \cellcolor{orange}0.312 & \cellcolor{orange}24.18 & \cellcolor{orange}0.775 & \cellcolor{yellow}0.222 \\
        Ours* & \cellcolor{red}20.51 & \cellcolor{red}0.625 & \cellcolor{red}0.309 & \cellcolor{red}24.25 & \cellcolor{red}0.783 & \cellcolor{red}0.206 \\
        \bottomrule
        \end{tabular}
    }
\end{table}

\begin{figure}[t] 
    \vspace{0mm}
    \centering
       \includegraphics[width=1.0\linewidth]{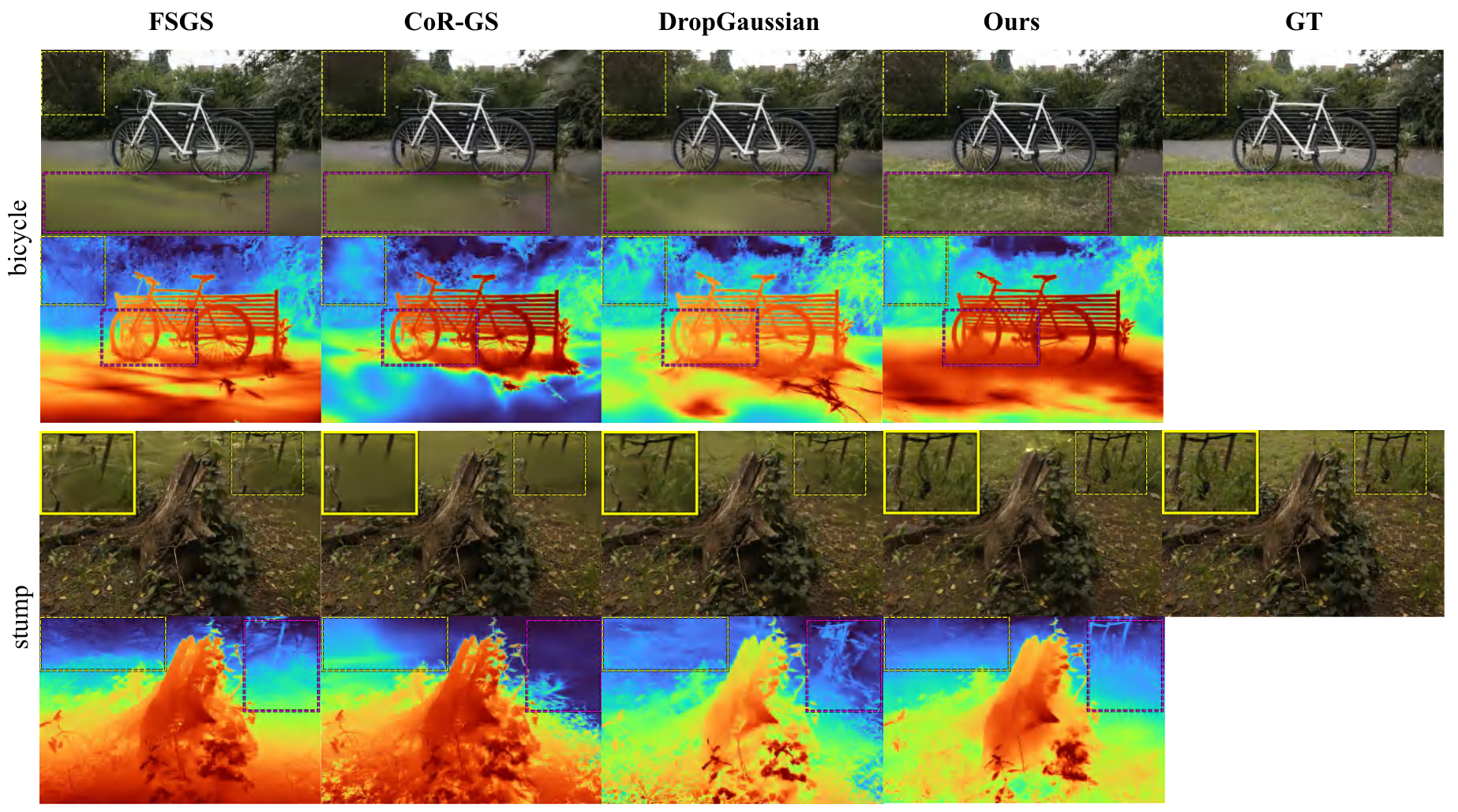}
       \caption{\textbf{Visual comparison on the Mip-NeRF360 dataset.}}
    \label{fig:visual_compare_MipNeRF360}
\end{figure}

\noindent\textbf{Comparison on DTU.}
Table \ref{tab:dtu} presents the quantitative results on the DTU dataset with 3, 6, and 9 input views. Our StereoGS achieves state-of-the-art performance across most evaluation metrics. Specifically, under the highly constrained 3-view setting, StereoGS demonstrates significant improvements in PSNR, SSIM, and LPIPS compared to recent approaches such as Binocular3DGS~\cite{han2024binocular} and MVPGS~\cite{xu2024mvpgs}. Fig. \ref{fig:visual_compare_DTU} provides visual comparisons for three scenes from the DTU dataset. As observed in the rendered novel views, DNGaussian~\cite{li2024dngaussian}, FSGS~\cite{zhu2023FSGS}, and MVPGS~\cite{xu2024mvpgs} yield blurry results, whereas Binocular3DGS~\cite{han2024binocular} exhibits numerous artifacts in textureless regions due to self-supervised photometric constraints. In contrast, our StereoGS significantly reduces artifacts in textureless regions while achieving superior rendering quality.

\noindent\textbf{Comparison on Mip-NeRF360.}
Table \ref{tab:mipnerf360} presents the quantitative results on the Mip-NeRF360 dataset using 12 and 24 input views. The proposed method achieves superior performance across most evaluation metrics. Specifically, when compared to recent state-of-the-art approaches, such as D$^2$GS~\cite{song2025Depth_and_Density} and DropGaussian~\cite{park2025dropgaussian}, StereoGS demonstrates significant improvements under both the 12-view and 24-view configurations. Figure \ref{tab:mipnerf360} provides visual comparisons of novel view synthesis and depth rendering for the \textit{bicycle} and \textit{stump} scenes from the Mip-NeRF360 dataset. In the \textit{bicycle} scene, baseline methods including FSGS~\cite{zhu2023FSGS}, CoR-GS~\cite{zhang2024cor-gs}, and DropGaussian~\cite{park2025dropgaussian} exhibit prominent artifacts in large textureless regions (for example, the grass areas indicated by dashed boxes), whereas the proposed approach maintains high rendering fidelity. Furthermore, in the \textit{stump} scene, these baseline approaches show difficulty in reconstructing fine-grained object details, which are successfully preserved by the proposed method. Consistent with the results of RGB rendering, the proposed method produces significantly more accurate and coherent depth rendering than the baselines.

\begin{figure}[t] 
    \vspace{0mm}
    \centering
       \includegraphics[width=\linewidth]{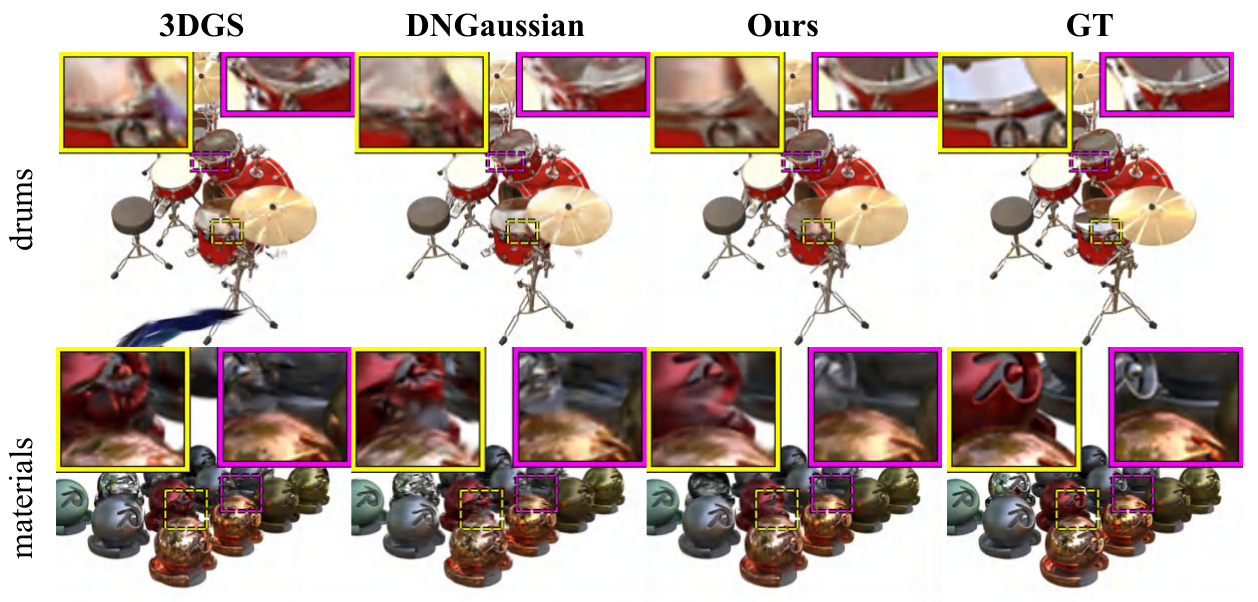}
       \caption{\textbf{Visualization on Blender under 1/2 resolution with 8 input views.}}
    \label{fig:vis_blender}
\end{figure}

\noindent\textbf{Comparison on Blender.}
Table \ref{tab:blender} presents the quantitative results on the Blender dataset using 8 input views. Our proposed method consistently outperforms all baselines across all metrics, demonstrating its superior capability in sparse-view reconstruction. Fig. \ref{fig:vis_blender} provides visual comparisons for two scenes from the Blender dataset.

\begin{table}[!t]
  \caption{
        \textbf{Quantitative results on Blender with 8 training views.} 
    }
  \label{tab:blender} 
  \renewcommand{\arraystretch}{0.8} % 调整行间距
  \centering
  \resizebox{0.7\linewidth}{!}{
      \setlength{\tabcolsep}{4mm}
      \begin{tabular}{l c c c} 
        \toprule
        Methods
        & PSNR$\uparrow$
        & SSIM$\uparrow$
        & LPIPS$\downarrow$ \\ 
    
        \midrule
        RegNeRF~\cite{niemeyer2022regnerf} & 23.86 & 0.852 & 0.105 \\
        FreeNeRF~\cite{yang2022freenerf} & 24.26 & 0.883 & 0.098 \\
        SparseNeRF~\cite{wang2023sparsenerf} & 24.04 & 0.876 & 0.113 \\
        
        \midrule
        3DGS~\cite{kerbl20233dgs} & 23.20 & 0.870 & 0.104 \\
        DNGaussian~\cite{li2024dngaussian} & 24.31 & 0.886 & 0.088 \\
        FSGS~\cite{zhu2023FSGS} & 24.64 & \cellcolor{yellow}0.895 & 0.095 \\
        CoR-GS~\cite{zhang2024cor-gs} & 24.43 & \cellcolor{orange}0.896 & \cellcolor{yellow}0.084 \\
        NexusGS~\cite{zheng2025nexusgs} & 24.37 & 0.893 & 0.087 \\ 
        DropGaussian~\cite{park2025dropgaussian} & 24.42 & 0.888 & 0.089 \\
        Binocular3DGS~\cite{han2024binocular} & \cellcolor{yellow}24.71 & 0.872 & 0.101 \\
        Ours & \cellcolor{orange}24.83 & \cellcolor{red}0.899 & \cellcolor{orange}0.081 \\
        Ours* & \cellcolor{red}25.04 & \cellcolor{red}0.899 & \cellcolor{red}0.078 \\
        \bottomrule
      \end{tabular}
    }
\end{table}

\begin{table}[h]
    \scriptsize
    \caption{\textbf{Ablation studies on LLFF and DTU datasets with 3 input views.}
    }
    \vspace{0mm}
    \centering
    \resizebox{\linewidth}{!}
    {
    \begin{tabular}{ccccccccc}
    \toprule

        \multirow{2}{2.5cm}{Consistency-Aware \\ Dense Initialization}
        & \multirow{2}{2cm}{Stereo Depth Regularization}
        & \multirow{2}{2cm}{Gradient-Aware Opacity Decay}
        & \multicolumn{3}{c}{LLFF}
        & \multicolumn{3}{c}{DTU} \\
        
        ~ & ~ & ~ 
        & PSNR$\uparrow$
        & SSIM$\uparrow$
        & LPIPS$\downarrow$
        & PSNR$\uparrow$
        & SSIM$\uparrow$
        & LPIPS$\downarrow$ \\
        
    \midrule
        \xmark & \xmark & \xmark  & 16.02 & 0.465 & 0.378 & 10.99 & 0.585 & 0.313 \\
        \cmark & \xmark & \xmark  & 19.75 & 0.691 & 0.215 & 14.10 & 0.786 & 0.196 \\
        \xmark & \cmark & \xmark  & 17.32 & 0.524 & 0.317 & 12.46 & 0.697 & 0.208 \\
        \xmark & \xmark & \cmark  & 18.18 & 0.569 & 0.291 & 15.05 & 0.751 & 0.202 \\
        \xmark & \cmark & \cmark  & 18.96 & 0.605 & 0.262 & \cellcolor{yellow}17.66 & 0.784 & 0.172 \\
        \cmark & \cmark & \xmark  & \cellcolor{yellow}19.79 & \cellcolor{yellow}0.695 & \cellcolor{yellow}0.214 & 15.57 & \cellcolor{yellow}0.812 & \cellcolor{yellow}0.166 \\
        \cmark & \xmark & \cmark  & \cellcolor{orange}21.18 & \cellcolor{orange}0.741 & \cellcolor{orange}0.171 & \cellcolor{orange}19.76 & \cellcolor{orange}0.863 & \cellcolor{orange}0.112 \\
        \cmark & \cmark & \cmark  & \cellcolor{red}21.91 & \cellcolor{red}0.773 & \cellcolor{red}0.157 & \cellcolor{red}21.46 & \cellcolor{red}0.879 & \cellcolor{red}0.099 \\

    \bottomrule
    \end{tabular}
    }
    \vspace{0mm}
    \label{table:ablation_llff_dtu}
\end{table}

\begin{figure}[t] 
    \vspace{0mm}
    \centering
       \includegraphics[width=\linewidth]{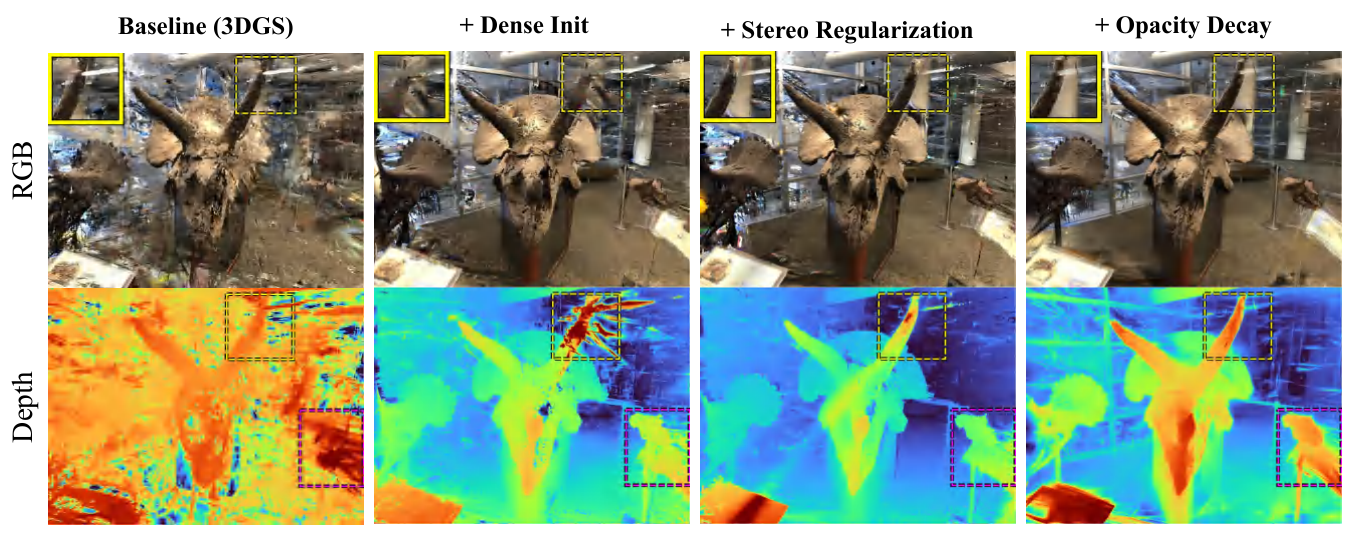}
       \caption{\textbf{Ablation Study of Visual Quality.} We compare the rendered RGB images and depth maps of novel views to verify the contributions of each proposed component.}
    \label{fig:ablation_llff_dtu}
\end{figure}

\subsection{Ablation Studies}

To verify the effectiveness of the proposed consistency-aware dense initialization, stereo depth regularization, and gradient-aware opacity decay strategy, we conduct ablation studies by incrementally integrating these components on the LLFF and DTU datasets. As demonstrated in Table \ref{table:ablation_llff_dtu}, the removal of any proposed module leads to a clear degradation in performance. 

\noindent\textbf{Effectiveness of Consistency-Aware Dense Initialization.} As shown in the second row of Table \ref{table:ablation_llff_dtu}, incorporating consistency-aware dense initialization substantially improves performance over the baseline (\eg, PSNR increases from 16.02 to 19.75 on the LLFF dataset and from 10.99 to 14.10 on the DTU dataset). As illustrated in Fig.~\ref{fig:ablation_llff_dtu}, the baseline 3DGS fails to reconstruct meaningful geometry, resulting in disordered depth maps and severe floaters. In contrast, the proposed dense initialization enables the clear recovery of the overall scene structure. This demonstrates that zero-shot MVS priors provide a more robust and crucial geometric foundation than sparse SfM points. Furthermore, to validate specific design choices, we conduct additional ablation studies using different MVS models and compare the generated point clouds with recent state-of-the-art methods in the supplementary material. However, despite this strong foundation, local details in the rendered depth maps, such as the dinosaur horn (indicated by the yellow dashed box in Fig.~\ref{fig:ablation_llff_dtu}), still contain severe holes and artifacts. This limitation necessitates the introduction of further geometric constraints.

\noindent\textbf{Effectiveness of Stereo Depth Regularization.} Building upon the dense initialization, the integration of stereo depth regularization further refines the geometry. Table~\ref{table:ablation_llff_dtu} demonstrates consistent quantitative improvements across all metrics. More importantly, Fig.~\ref{fig:ablation_llff_dtu} illustrates the substantial qualitative impact of this module: severe depth holes on the horn are effectively filled, and the background geometry becomes significantly smoother. To verify that these improvements arise from the proposed regularization formulation rather than a specific network architecture, we evaluate the results of different stereo matching models in the supplementary material. 
% These results confirm that the pseudo-ground-truth stereo depth successfully mitigates scale ambiguity and provides reliable geometric constraints, thereby preventing the Gaussians from overfitting to sparse training views.

\noindent\textbf{Effectiveness of Gradient-Aware Opacity Decay.} 
Finally, the replacement of the standard periodic reset with the proposed gradient-aware opacity decay strategy yields the most significant visual and quantitative refinements. This module achieves a best performance of 21.91 PSNR on the LLFF dataset and 21.46 PSNR on the DTU dataset, and it excels at pruning redundant noise. As shown in the final column of Fig.~\ref{fig:ablation_llff_dtu}, the remaining floaters in the background and around the boundaries of the foreground are cleanly eliminated. By dynamically evaluating the importance of each Gaussian through the relative gradient, the proposed strategy successfully preserves essential surface structures while pruning noise points. To rigorously justify our opacity decay formulation, we compare our exponential decay function against alternative mapping strategies (\eg, constant, step, and linear decay) and raw gradient inputs in the supplementary material.

\section{Conclusion}
\label{sec:conclusion}

In this paper, we presented StereoGS, a novel framework that effectively addresses the critical challenges of scale ambiguity and geometric inconsistency in sparse-view 3D Gaussian Splatting. 
By seamlessly integrating foundation stereo models, we introduced a Stereo Depth Regularization mechanism that constructs virtual stereo pairs to enforce absolute scale and binocular consistency during optimization. 
Furthermore, we designed a Gradient-Aware Opacity Decay strategy that dynamically prunes redundant primitives based on their relative optimization gradients, preserving essential geometry while eliminating noise. 
Coupled with a Consistency-Aware Dense Initialization, StereoGS establishes a robust geometric foundation from the outset. 
Extensive experiments across multiple standard benchmarks demonstrate that StereoGS achieves state-of-the-art novel view synthesis quality under sparse-view conditions, all while maintaining the real-time rendering efficiency of vanilla 3DGS.

\noindent\textbf{Limitations.} Stereo models may produce inaccurate predictions in extremely textureless regions. Furthermore, online stereo depth regularization inevitably incurs additional training overhead. Addressing these challenges represents a promising direction for future research.

\bibliographystyle{splncs04}
\bibliography{main}

\newpage
\begin{center}
    \textbf{\large StereoGS: Sparse-View 3D Gaussian Splatting via Stereo Priors}
\end{center}
\begin{center}
    \textbf{\large Supplementary Material}
\end{center}

\appendix

\section{Role of Opacity Gradient}

We use opacity gradient only as an optimization-state indicator, not as a measure of geometric importance. Geometric consistency is enforced by Stereo Depth Regularization, and the opacity gradient is computed from the total loss, including the stereo depth loss. Thus, depth inconsistency also affects the decay signal, making the opacity decay complement stereo regularization.
For a floater, a large opacity gradient does not freeze or increase its opacity. It only reduces the extra decay at the current step. The floater is still updated by the photometric and depth losses. If it remains unsupported after optimization, its opacity will decrease and be pruned.

% \section{Training Overhead \& 3DGS Count}
% As shown in Table~\ref{tab:training_overhead}, although our method increases training time, this is an offline cost. StereoGS uses less peak VRAM and produces significantly fewer Gaussians.
% \begin{table}[htbp]
%   \vspace{0mm}
%   \caption{\textbf{Average Training Overhead and 3DGS Count.}}
%   \label{tab:training_overhead} 
%   \centering
%   \scriptsize
%   % \vspace{1mm} % 标题与表格的微小呼吸空间
%   % \renewcommand{\arraystretch}{0.5} % 调整行间距，使表格更紧凑
%   \resizebox{\linewidth}{!}{ % 自动缩放至当前栏宽
%       % \setlength{\tabcolsep}{1.5mm} % 稍微缩小列间距以适应 7 列
%       \begin{tabular}{l c c c c c c} 
%         \toprule
%         \multirow{2}{*}[-0.5ex]{Methods}
%         & \multicolumn{2}{c}{Train Time (mins)}
%         & \multicolumn{2}{c}{Peak VRAM (GB)} 
%         & \multicolumn{2}{c}{3DGS Count} \\ 
%         \cmidrule(lr){2-3} \cmidrule(lr){4-5} \cmidrule(l){6-7}

%         & \multicolumn{1}{c}{LLFF (3-view)}
%         & \multicolumn{1}{c}{DTU (3-view)}
%         & \multicolumn{1}{c}{LLFF (3-view)}
%         & \multicolumn{1}{c}{DTU (3-view)} 
%         & \multicolumn{1}{c}{LLFF (3-view)}
%         & \multicolumn{1}{c}{DTU (3-view)} \\ 
    
%         \midrule
%         3DGS & 4.3 & 3.9 & 0.5 & 0.3 & 265.1K & 168.6K \\
%         FSGS & 26.5 & 15.6 & 11.8 & 3.7 & 370K & 71.6K \\
%         MVPGS & 6.4 & 5.8 & 5.2 & 4.8 & 1521.7K & 220.9K \\
%         \textbf{StereoGS (Ours)} & 43.2 & 34.6 & \textbf{3.6} & \textbf{2.5} & \textbf{126K} & \textbf{44.4K} \\
%         \bottomrule
%         \end{tabular}
%     }
%   \vspace{0mm} % 消除表格下方的多余留白
% \end{table}

\section{Failure Cases}
\begin{figure}[htbp]
    \centering
    % \vspace{-4mm}
    \includegraphics[width=\linewidth]{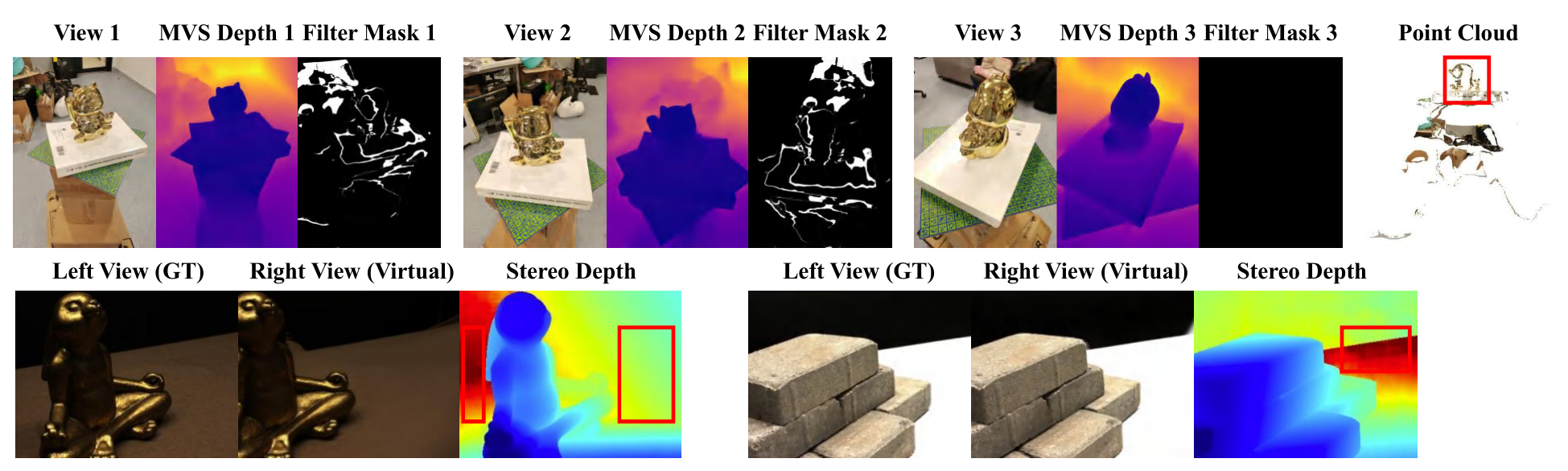}
    % \vspace{-4mm}
    \caption{\textbf{Visualization of failure cases} }
    % \vspace{-5mm}
    \label{fig:failure_cases}
\end{figure}

Fig.~\ref{fig:failure_cases} shows two cases. First, Consistency-Aware Dense Initialization may fail on reflective objects on specific view, as view-dependent reflections make MVS depths filter by reprojection filtering. Second, large textureless regions can make stereo matching ambiguous, producing erroneous stereo depth and inaccurate geometric constraints.

% \subsection{Limitations}
% First, MVS-based initialization can fail in some views for highly reflective or textureless objects, leading to incomplete initial geometry. Second, the stereo model still estimates reasonable depth on the target object, but it may produce inaccurate predictions in weakly textured background floor. Improving robustness to imperfect initialization and stereo priors is an important direction. 
\section{Additional Ablation Results}

\subsection{Ablation Study on Different MVS Models}

To validate the design choice of our Consistency-Aware Dense Initialization module, we compare different Multi-View Stereo (MVS) models for point cloud initialization. As discussed in the main paper, relying solely on sparse point clouds generated by Structure-from-Motion (SfM)~\cite{schonberger2016structure} provides insufficient geometric guidance under sparse-view settings. This inevitably leads to suboptimal novel view synthesis, yielding only 18.96 dB and 17.66 dB in PSNR on the LLFF and DTU datasets, respectively, under the 3-view setting.

To address this, we evaluate the integration of various learning-based MVS priors, including PDCNet+~\cite{truong2023pdc}, MVSFormer~\cite{cao2022mvsformer}, and our adopted MVSAnywhere~\cite{izquierdo2025mvsanywhere}. As detailed in \cref{tab:mvs_ablation_llff_dtu}, while PDCNet+ and MVSFormer improve upon the SfM baseline, their performance is constrained by the domain gap between their pre-training data and diverse in-the-wild scenes. In contrast, MVSAnywhere leverages powerful zero-shot generalization to extract highly accurate, multi-view-consistent depth maps. Consequently, StereoGS equipped with MVSAnywhere achieves the best quantitative results across all metrics.

\begin{figure}[h]
    \centering
    \includegraphics[width=\linewidth]{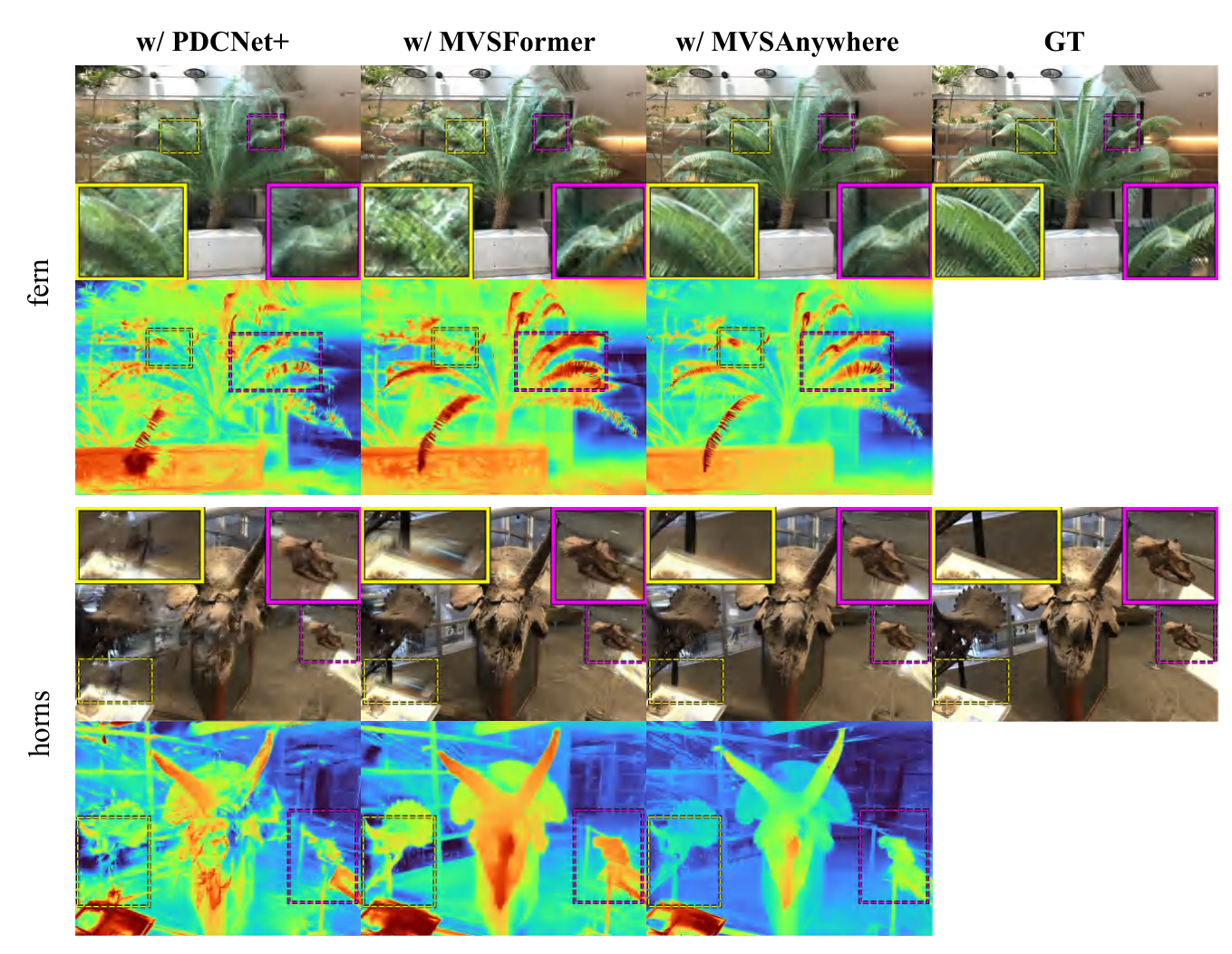}
    \caption{\textbf{Qualitative comparison of different MVS initialization models.} Visual results on the \textit{fern} and \textit{horns} scenes demonstrate that while our framework benefits from various MVS priors, the default configuration (with MVSAnywhere) synthesizes the sharpest details and the most accurate depth maps compared to other variants.}
    \label{fig:ablation_mvs}
\end{figure}

This superiority is further corroborated by the qualitative comparisons in \cref{fig:ablation_mvs}. Visualizations of the \textit{fern} and \textit{horns} scenes demonstrate that models initialized with PDCNet+ or MVSFormer still struggle with geometric artifacts and incomplete structures in challenging regions. Conversely, MVSAnywhere provides a significantly more robust and dense geometric foundation, enabling our framework to render sharper details and highly accurate depth maps that closely align with the ground truth (GT).

\begin{table}[h]
    \scriptsize
    \caption{\textbf{Quantitative results of different MVS models on LLFF and DTU datasets with 3 input views.}}
    \vspace{0mm}
    \centering
    \resizebox{0.9\linewidth}{!}
    {
        \begin{tabular}{lcccccc}
        \toprule
            \multirow{2}{*}{Methods}
            & \multicolumn{3}{c}{LLFF}
            & \multicolumn{3}{c}{DTU} \\
            ~ 
            & PSNR$\uparrow$
            & SSIM$\uparrow$
            & LPIPS$\downarrow$
            & PSNR$\uparrow$
            & SSIM$\uparrow$
            & LPIPS$\downarrow$ \\
            
        \midrule
            Ours (w/ SfM~\cite{schonberger2016structure}) & 18.96 & 0.605 & 0.262 & 17.66 & 0.784 & 0.172 \\
            Ours (w/ PDCNet+~\cite{truong2023pdc}) & 20.10 & 0.666 & 0.222 & 19.81 & 0.843 & 0.133 \\
            Ours (w/ MVSFormer~\cite{cao2022mvsformer}) & 21.08 & 0.737 & 0.176 & 20.71 & 0.865 & 0.115 \\
            \textbf{Ours (w/ MVSAnywhere~\cite{izquierdo2025mvsanywhere})} & \textbf{21.91} & \textbf{0.773} & \textbf{0.157} & \textbf{21.46} & \textbf{0.879} & \textbf{0.099} \\ 
            
        \bottomrule
        \end{tabular}
    }
    \vspace{0mm}
    \label{tab:mvs_ablation_llff_dtu}
\end{table}

\subsection{Ablation Study on Different Stereo Models}

To demonstrate the universality and adaptability of the proposed Stereo Depth Regularization, we evaluate the compatibility of the module with different stereo matching priors. Specifically, we substitute the default FoundationStereo~\cite{wen2025foundationstereo} prior with other recent large-scale generalized stereo models, including S2M2~\cite{min2025s2m2} and LiteAnyStereo~\cite{jing2025lite}. To ensure a fair comparison, we replace only the stereo matching model, keeping all other network components and training settings unchanged.

As reported in Table~\ref{tab:stereo_ablation}, the integration of any of these stereo priors yields consistent and significant improvements over the baseline. This observation indicates that the proposed depth regularization strategy is not limited to a specific stereo architecture. Instead, the strategy serves as a universal component that effectively leverages general stereo priors to enhance novel view synthesis.

\begin{figure}[h]
    \centering
    \includegraphics[width=\linewidth]{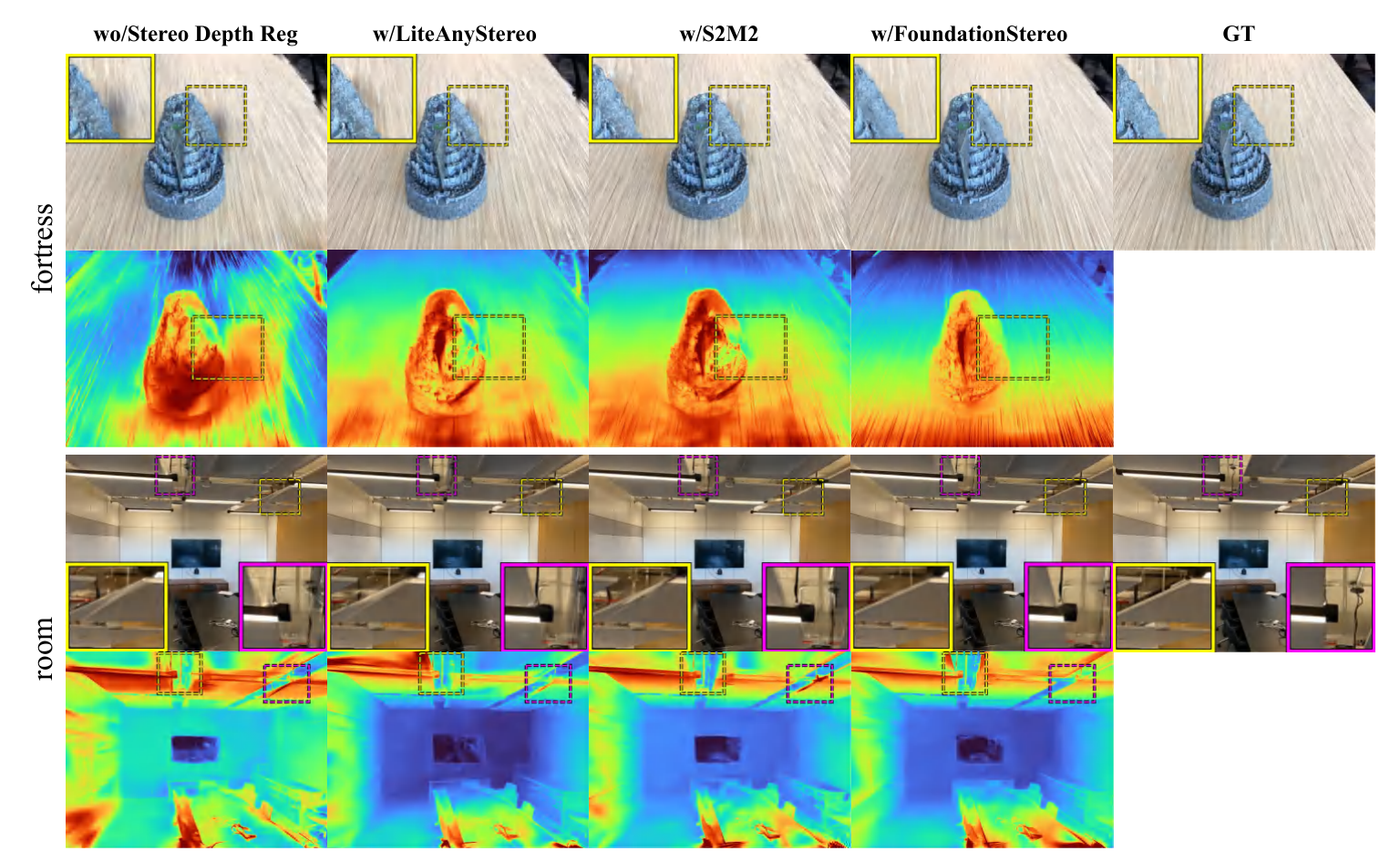}
    \caption{\textbf{Qualitative comparison of different stereo models.} Visual results on the \textit{fortress} and \textit{room} scenes demonstrate that while the proposed regularization module is compatible with various priors, the default configuration (with FoundationStereo) synthesizes the sharpest details and the fewest artifacts compared to other variants and the baseline.}
    \label{fig:stereo_ablation}
\end{figure}

Among the used models, the default configuration using FoundationStereo achieves the most substantial performance gains. As further illustrated in Fig.~\ref{fig:stereo_ablation}, qualitative comparisons on the \textit{fortress} and \textit{room} scenes demonstrate that the proposed method equipped with FoundationStereo produces the sharpest geometric details and the fewest rendering artifacts. This result justifies the selection of FoundationStereo as the default prior while confirming the overall flexibility and generalizability of the proposed framework.

We provide the qualitative comparisons in Fig.~\ref{fig:stereo_ablation}. The visualizations verify that using any stereo model improves detail, and FoundationStereo yields the sharpest geometric details and the fewest rendering artifacts. This confirms the overall flexibility and generalizability of the proposed framework.

\begin{table}[h]
    \scriptsize
    \caption{\textbf{Quantitative results of different stereo models on LLFF and DTU datasets with three input views.}}
    \vspace{0mm}
    \centering
    \resizebox{0.9\linewidth}{!}
    {
        \begin{tabular}{lcccccc}
        \toprule
            \multirow{2}{*}{Methods}
            & \multicolumn{3}{c}{LLFF}
            & \multicolumn{3}{c}{DTU} \\
            ~ 
            & PSNR$\uparrow$
            & SSIM$\uparrow$
            & LPIPS$\downarrow$
            & PSNR$\uparrow$
            & SSIM$\uparrow$
            & LPIPS$\downarrow$ \\
            
        \midrule
            Ours (w/o Stereo Depth Reg) & 21.18 & 0.741 & 0.171 & 19.76 & 0.863 & 0.112  \\
            Ours (w/ LiteAnyStereo~\cite{jing2025lite}) & 21.28 & 0.747 & 0.166 & 20.84 & 0.841 & 0.111 \\
            Ours (w/ S2M2~\cite{min2025s2m2}) & 21.53 & 0.758 & 0.163 & 21.12 & 0.854 & 0.108 \\
            \textbf{Ours (w/ FoundationStereo~\cite{wen2025foundationstereo})} & \textbf{21.91} & \textbf{0.773} & \textbf{0.157} & \textbf{21.46} & \textbf{0.879} & \textbf{0.099} \\ 
            
        \bottomrule
        \end{tabular}
    }
    \vspace{0mm}
    \label{tab:stereo_ablation}
\end{table}

\subsection{Ablation Study on the Validity Mask}
To validate the necessity of the validity mask ($M_{\text{valid}}$) in our Stereo Depth Regularization, we conduct an ablation study on the LLFF and DTU datasets with 3 input views. Specifically, we remove the mask during optimization and apply the stereo depth loss to all pixels indiscriminately, keeping all other components and training settings entirely unchanged.

As reported in Table~\ref{tab:mask_ablation}, removing the validity mask leads to a noticeable performance drop across both datasets. This decline occurs because foundation stereo models inevitably produce erroneous disparity estimates in occluded, textureless, or anomalous regions. Without the validity mask to filter out these unreliable priors, the optimization process is misguided by noisy depth supervision, thereby degrading the overall rendering quality.

\begin{figure}[t] 
    \vspace{0mm}
    \centering
       \includegraphics[width=0.8\linewidth]{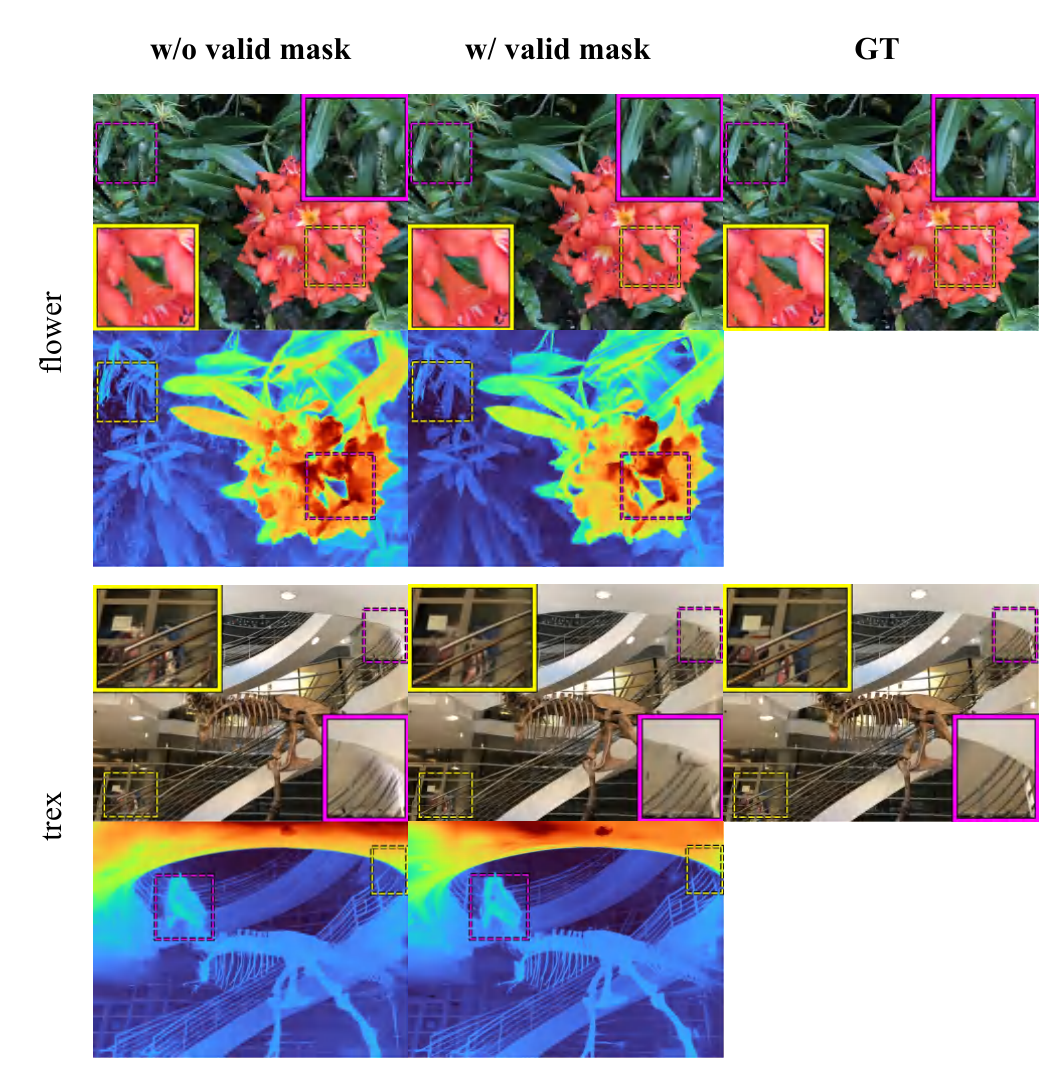}
       \caption{\textbf{Qualitative comparison of the validity mask on the LLFF dataset with 3 input views.} Visual results on the \textit{flower} and \textit{trex} scenes demonstrate that optimizing without the validity mask leads to noticeable floaters and geometric distortions around complex boundaries. By filtering out unreliable stereo priors, our model (w/ $M_{\text{valid}}$) synthesizes significantly cleaner and more accurate geometry.}
    \label{fig:ablation_valid_mask}
\end{figure}

Qualitative comparisons on the \textit{flower} and \textit{trex} scenes in Fig.~\ref{fig:ablation_valid_mask} further support this conclusion. As shown in the visualizations, optimizing without the validity mask introduces severe geometric artifacts and floaters, particularly around complex object boundaries and heavily occluded areas (e.g., the background of the flower and the skeletal structure of the trex). In contrast, our model (w/ $M_{\text{valid}}$), which leverages the validity mask to enforce supervision strictly on reliable regions, successfully preserves clean and accurate scene geometry.

\begin{table}[h]
    \scriptsize
    \caption{\textbf{Quantitative results of the validity mask ablation on the LLFF and DTU datasets with 3 input views.}}
    \vspace{0mm}
    \centering
    \resizebox{0.8\linewidth}{!}
    {
        \begin{tabular}{lcccccc}
        \toprule
            \multirow{2}{*}{Methods}
            & \multicolumn{3}{c}{LLFF}
            & \multicolumn{3}{c}{DTU} \\
            ~ 
            & PSNR$\uparrow$
            & SSIM$\uparrow$
            & LPIPS$\downarrow$
            & PSNR$\uparrow$
            & SSIM$\uparrow$
            & LPIPS$\downarrow$ \\
            
        \midrule
            Ours (w/o $M_{\text{valid}}$) & 21.30 & 0.748 & 0.165 & 20.82 & 0.841 & 0.107 \\
            \textbf{Ours (w/ $M_{\text{valid}}$)} & \textbf{21.91} & \textbf{0.773} & \textbf{0.157} & \textbf{21.46} & \textbf{0.879} & \textbf{0.099} \\ 
            
        \bottomrule
        \end{tabular}
    }
    \vspace{0mm}
    \label{tab:mask_ablation}
\end{table}

\subsection{Ablation Study on Opacity Decay Strategies}

\begin{figure}[t] 
    \vspace{0mm}
    \centering
       \includegraphics[width=1.0\linewidth]{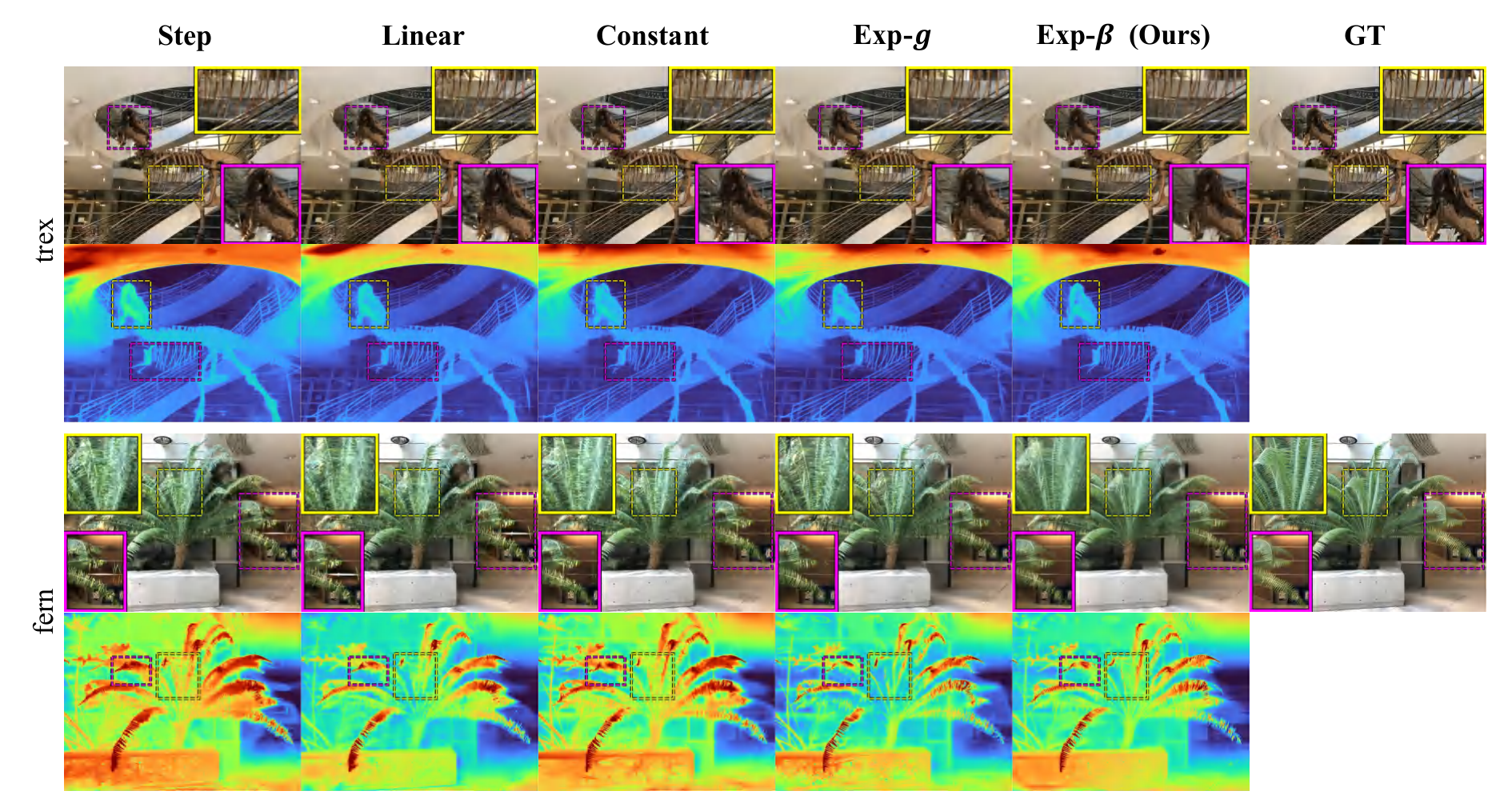}
       \caption{\textbf{Visual comparison of different opacity decay strategies on the LLFF dataset with 3 input views.}}
    \label{fig:ablation_opacity}
\end{figure}

\begin{figure}[t]
    \centering
    \begin{minipage}{0.48\textwidth}
        \centering
        \includegraphics[width=\linewidth]{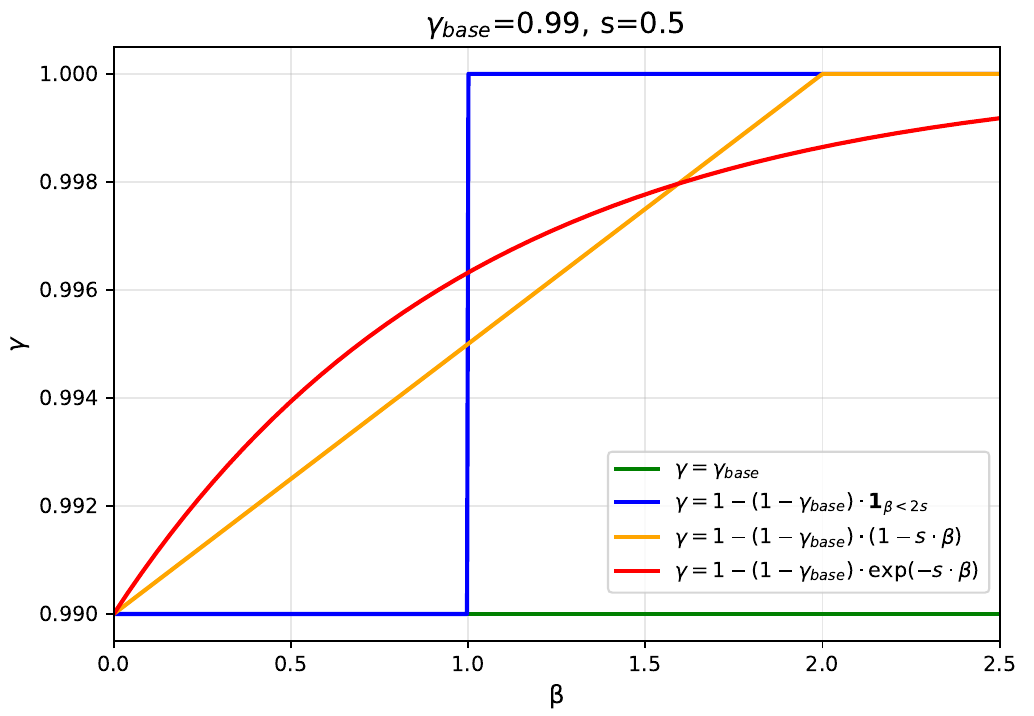}
        \caption{\textbf{Visualization of Opacity Decay Functions.}}
        \label{fig:curve_opacity_ablation}
    \end{minipage}\hfill
    \begin{minipage}{0.48\textwidth}
        \centering
        \includegraphics[width=\linewidth]{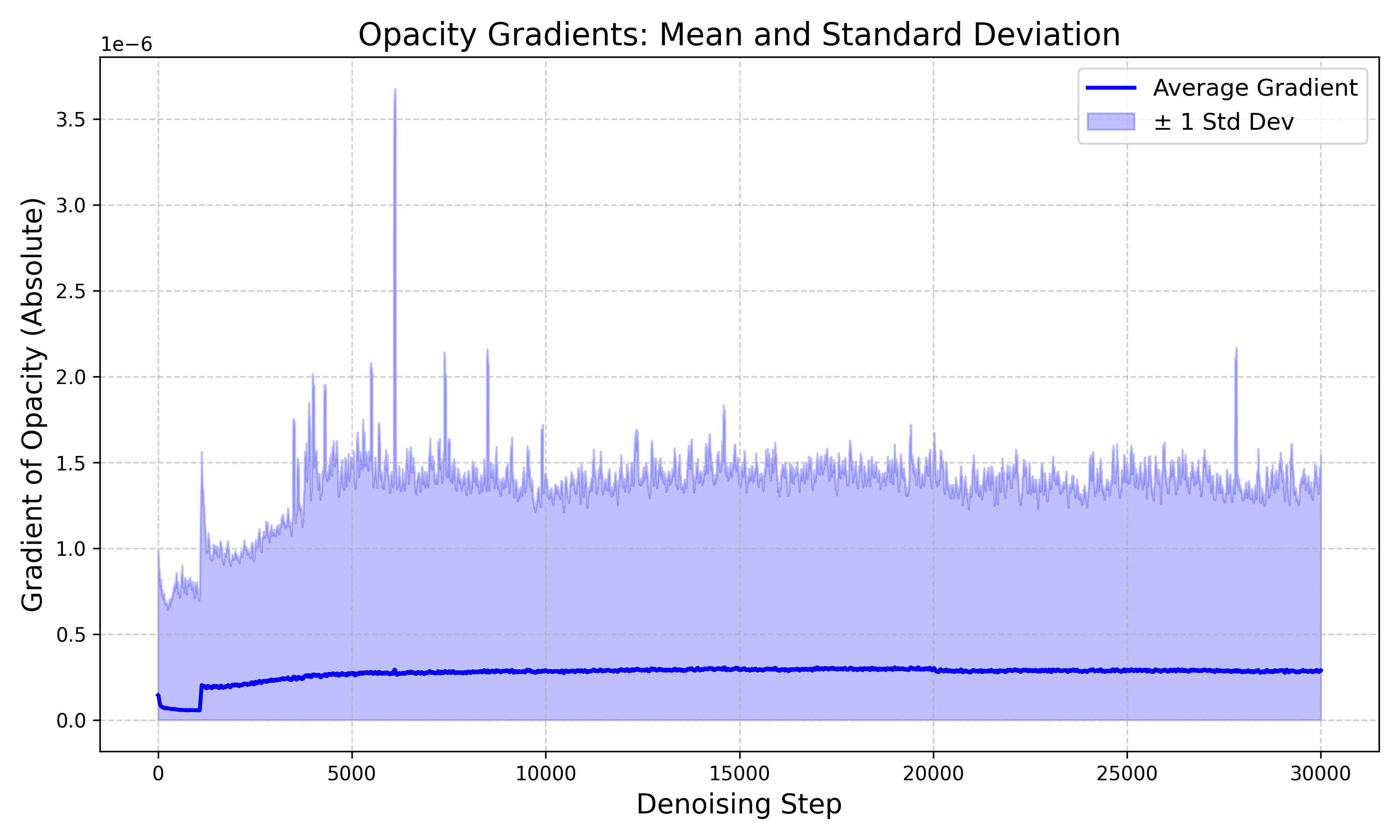}
        \caption{\textbf{Statistical distribution of opacity gradients during training.}}
        \label{fig:opacity_gradients_std}
    \end{minipage}
\end{figure}

To validate the design of the proposed Gradient-Aware Opacity Decay module, we compare the default exponential strategy against several other functions: Constant, Step, and Linear.
Table~\ref{table:ablation_opacity_decay} presents the quantitative ablation results on the LLFF dataset. 
Fig.~\ref{fig:curve_opacity_ablation} provides the curves of different functions for intuitive understanding. Fig.~\ref{fig:opacity_gradients_std} illustrates the statistical distribution of opacity gradients during training. Fig.~\ref{fig:ablation_opacity} provides qualitative visual results.

We first analyze the Constant strategy~\cite{han2024binocular}, which applies a uniform decay penalty across all Gaussians. The primary limitation of this approach is the lack of discrimination: it fails to distinguish between important Gaussians and floaters. As shown in the highlighted regions of Fig.~\ref{fig:ablation_opacity}, this indiscriminate penalty incorrectly suppresses essential geometric structures, leading to rendering artifacts in the RGB images and missing surfaces in the depth maps.

To address this limitation, we ablate a Step or Linear function based on the relative gradient $\beta$. However, as shown in Table~\ref{table:ablation_opacity_decay}, these functions yield lower PSNR scores (21.17 and 21.24) than the Constant baseline. The mathematical curves in Fig.~\ref{fig:curve_opacity_ablation} explain this failure: since the majority of gradients exceed the mean value (i.e., $\beta>1$), both Step and Linear functions tend to maintain high opacity for an excessive number of invalid Gaussian primitives. Specifically, the Step function causes abrupt truncations that fail to filter these regions, while the Linear function insufficiently penalizes primitives with moderate-to-high gradients. This leads to a suboptimal quantitative performance compared to the Constant decay.

\noindent \textbf{Raw or Relative Gradient Norm.} As illustrated in Fig.~\ref{fig:opacity_gradients_std}, the absolute opacity gradients ($g$) during training are extremely small, predominantly on the order of $10^{-6}$. Consequently, when applying exponential decay directly to these raw gradients (Exp-$g$), the exponential term exp (Exp-$g$) remains nearly constant at approximately 1.0. This causes the decay function to degenerate into the Constant function, which is why its quantitative performance (21.41 PSNR) is close to the Constant results in Table~\ref{table:ablation_opacity_decay}. This observation underscores the necessity of our relative gradient $\beta$, which provides a properly scaled normalization to ensure stable and effective opacity attenuation.

Our default exponential function using the relative gradient $\beta$ (Exp-$\beta$) strategy serves as an effective solution, achieving a significant 21.91 PSNR. Inspired by the concept of relative advantage in GRPO~\cite{shao2024deepseekmath}, the relative metric $\beta$ normalizes the small and noisy absolute gradients. This normalization provides a stable, scale-invariant measure of structural importance. Exp-$\beta$ accurately targets and heavily penalizes below-average floaters ($\beta < 1$) to remove background noise, while it rapidly increases the retention rate for above-average valid structures. As shown in Fig.~\ref{fig:ablation_opacity}, this collaboration effectively preserves high-frequency details, yielding sharp novel-view synthesis and complete depth maps.

\begin{table}[t]
    \caption{
        \textbf{Quantitative results of different opacity decay strategies on LLFF with 3 input views.} 
    }
    \vspace{0mm}
    \centering
    \resizebox{0.9\textwidth}{!}{
    \begin{tabular}{l l l c c c}
        \toprule
            Independent Variable
            & Decay Strategies 
            & Function Formula 
            & PSNR$\uparrow$
            & SSIM$\uparrow$
            & LPIPS$\downarrow$ \\
            
        \midrule  
            ~ & Constant \cite{han2024binocular} & $\hat{\alpha} = \alpha \cdot  \gamma_{\text{base}}$ & 21.45 & 0.751 & 0.164  \\
            
            \midrule
            
            \multirow{3}{*}{$\beta=g/\bar{g}$} & Step & $\hat{\alpha} = \alpha \cdot (1 - (1 - \gamma_{\text{base}})\cdot \mathbf{1}_{\beta<2s} )$ & 21.17 & 0.747 & 0.163  \\ 
            
            ~ & Linear & $\hat{\alpha} = \alpha \cdot (1 - (1 - \gamma_{\text{base}}) \cdot (1-s \cdot \beta))$ & 21.24 & 0.747 & 0.164  \\

            ~ & \textbf{Exp-$\beta$ (Ours)} & $\hat{\alpha} = \alpha \cdot (1 - (1 - \gamma_{\text{base}}) \cdot\exp(-s \cdot \beta))$ & \textbf{21.91} & \textbf{0.773} & \textbf{0.157}  \\
            \midrule

            $g$ & Exp-$g$ & $\hat{\alpha} = \alpha \cdot (1 - (1 - \gamma_{\text{base}}) \cdot \exp(-s \cdot g ))$ & 21.41 & 0.742 & 0.171 \\
            
        \bottomrule
        \end{tabular}
    }
    \vspace{-4mm}
    \label{table:ablation_opacity_decay}
\end{table}

\subsection{Ablation study for hyperparameter $\gamma_{\text{base}}$}
We perform ablation studies on the LLFF and DTU datasets to investigate the impact of the hyperparameters in our Gradient-Aware Opacity Decay strategy. We first evaluate the base decay factor $\gamma_{\text{base}}$, varying its value from 0.960 to 1.0 under the 3-view setting, with results presented in Table \ref{tab:ablation_gamma}. Note that setting $\gamma_{\text{base}} = 1.0$ is mathematically equivalent to disabling the opacity decay entirely. 

As shown in Table \ref{tab:ablation_gamma}, setting $\gamma_{\text{base}}$ too low (e.g., 0.960) leads to aggressive pruning, which mistakenly eliminates essential surface primitives and degrades the rendering quality. Conversely, a value too close to 1.0 fails to effectively suppress redundant floaters, leading to overfitting and artifacts. We find that the best balance between preserving stable structures and pruning noise is achieved when $\gamma_{\text{base}}$ is set to 0.99 on both datasets. 

Furthermore, we investigate the sensitivity hyperparameter $s$, which controls how easily a Gaussian can escape the opacity penalty based on its normalized gradient. According to our formulation, an excessively small $s$ (e.g., 0.1) dictates that a Gaussian requires a significantly large gradient to avoid the penalty, resulting in the indiscriminate pruning of useful primitives. On the other hand, an excessively large $s$ (e.g., 1.0) makes the mechanism overly sensitive, protecting even noisy floaters with minor gradients from being pruned and leaving redundant artifacts in the scene. Therefore, we adopt a balanced value of $s = 0.5$. 

\begin{table}[h]
    \caption{\textbf{Ablation studies for $\gamma_{\text{base}}$ on LLFF and DTU datasets with 3 input views.}}
    \centering
    \begin{tabular}{lllllllllll}
    \hline
    $\gamma_{\text{base}}$ & & 0.960 & 0.970 & 0.980 & 0.985 & 0.990 & 0.995 & 1.0 \\ \hline
    \multirow{3}{*}{LLFF} & PSNR$\uparrow$  & 19.12 & 20.45 & 21.25 & \cellcolor{orange}21.68 & \cellcolor{red}21.91 & \cellcolor{yellow}21.45 & 19.79 \\
    & SSIM$\uparrow$  & 0.612 & 0.695 & 0.745 & \cellcolor{orange}0.761 & \cellcolor{red}0.773 & \cellcolor{yellow}0.755 & 0.695 \\
    & LPIPS$\downarrow$ & 0.315 & 0.242 & 0.185 & \cellcolor{orange}0.168 & \cellcolor{red}0.157 & \cellcolor{yellow}0.165 & 0.214 \\ \hline
    \multirow{3}{*}{DTU}  & PSNR$\uparrow$  & 13.55 & 16.80 & 19.85 & \cellcolor{orange}20.85 & \cellcolor{red}21.46 & \cellcolor{yellow}20.60 & 15.57 \\
    & SSIM$\uparrow$  & 0.710 & 0.795 & 0.852 & \cellcolor{orange}0.865 & \cellcolor{red}0.879 & \cellcolor{yellow}0.868 & 0.812 \\
    & LPIPS$\downarrow$ & 0.250 & 0.185 & 0.125 & \cellcolor{orange}0.110 & \cellcolor{red}0.099 & \cellcolor{yellow}0.108 & 0.166 \\ \hline
    \end{tabular}
    \label{tab:ablation_gamma}
\end{table}

\subsection{Ablation Study for the Baseline $d$ between Virtual Camera Pairs.}

We provide the ablation results for the baseline $d$ between virtual left-right camera pairs, as shown in Table \ref{tab:ablation_d}. We use $d=4.0$ as our default setting.

\begin{table}[!h] % 使用 [!h] 强制固定位置，消除下方大段空白
    \caption{\textbf{Ablation studies for $d$ on LLFF with 3 input views.}}
    \label{tab:ablation_d}
    \tiny
    \centering
    % \scriptsize
    \renewcommand{\arraystretch}{0.75}
    \resizebox{0.9\linewidth}{!}{ % 强制缩放至单栏宽度，解决越界问题
        \begin{tabular}{c l c c c c c c c c c}
        \toprule
        % 使用 multicolumn 合并前两列，让 d 的数值与下方数据完美对齐
        \multicolumn{2}{c}{$d$} & 0.5 & 1.0 & 2.0 & 3.0 & 4.0 & 5.0 & 6.0 & 8.0 & 10.0 \\ 
        % \cline{1-11}
        \midrule
        \multirow{3}{*}{LLFF} 
        & PSNR$\uparrow$ & 21.71 & 21.71 & 21.86 & 21.89 & \textbf{21.91} & 21.90 & 21.88 & 21.86 & 21.85 \\
        & SSIM$\uparrow$ & 0.758 & 0.759 & 0.764 & 0.771 & \textbf{0.773} & 0.773 & 0.772 & 0.769 & 0.769 \\
        & LPIPS$\downarrow$ & 0.173 & 0.174 & 0.166 & 0.163 & \textbf{0.157} & 0.157 & 0.159 & 0.162 & 0.163 \\ 
        \bottomrule
        \end{tabular}
    }
\end{table}

\section{Experiment Details}

\subsection{Implementation Details}
To initialize the 3D Gaussians, we employ the pre-trained MVSAnywhere~\cite{izquierdo2025mvsanywhere} model to extract depth maps from the training views. These depth maps are subsequently fused into a dense point cloud via a geometric consistency filtering strategy~\cite{yao2018mvsnet}. During optimization, we densify the Gaussians every 100 iterations, beginning at iteration 1,000. For Stereo Depth Regularization, we adopt FoundationStereo~\cite{wen2025foundationstereo} as the stereo matching model. Within the Gradient-Aware Opacity Decay module, we set the hyperparameters to $\gamma_{\text{base}} = 0.99$ and $s = 0.5$. The training schedule varies across datasets. For the LLFF, DTU, and Mip-NeRF360 datasets, we train the model for a total of 30,000 iterations and apply Stereo Depth Regularization at iteration 20,000. For the Blender dataset, we set the total number of iterations to 7,000 and enable regularization at iteration 4,000. Furthermore, we report the performance of two variants of the proposed approach: the standard method without dropout (denoted as $Ours$) and a variant with a fixed dropout rate of 0.3 (denoted as $Ours^*$)~\cite{park2025dropgaussian}. All experiments are conducted on a single NVIDIA RTX 4090 GPU.

\subsection{Details of Valid Mask Generation in Stereo Depth Regularization}
As mentioned in Section 3.2 of the main text, the validity mask $M_{\text{valid}}$ is designed to eliminate occlusions, background, and matching anomalies. It is formulated as $M_{\text{valid}} = 1 - (M_{\text{bg}} \lor M_{\text{occ}} \lor M_{\text{anomaly}})$. The specific definitions are as follows:

\noindent\textbf{Left-Right Consistency Occlusion Mask} ($M_{\text{occ}}$). To identify occluded regions, we perform a left-right disparity consistency check. For a pixel coordinate $(u, v)$, the mask is defined as:
\begin{equation}
    M_{\text{occ}}(u, v) = 
    \begin{cases} 
        1, & \left|\hat{D}_{l}(u, v) - \hat{D}_{r}(u - \hat{D}_{l}(u, v), v)\right| > \tau \\ 
        0, & \text{otherwise} 
    \end{cases}
\end{equation}
where $\tau = 2.0$ is the consistency threshold.

\noindent\textbf{Background Mask} ($M_{\text{bg}}$). The strategy for generating the background mask varies slightly depending on the dataset characteristics to effectively filter out background regions:

For datasets with dark or unlit backgrounds (e.g., DTU), we extract the mask based on the RGB intensity of the training ground-truth image $I_{\text{gt}}$. If the maximum of R, G, B channels of a pixel is less than an intensity threshold $\theta_{\text{bg}}$, we consider this pixel as background. The mask is formulated as:
\begin{equation}
M_{\text{bg}}(u, v) = 
    \begin{cases} 
        1, & \max_{c \in \{R, G, B\}} I_{\text{gt}}^c(u, v) < \theta_{\text {bg}} \\ 
        0, & \text {otherwise} 
    \end{cases}
\end{equation}
    
For synthetic datasets that provide RGBA images (e.g., Blender), we directly utilize the ground-truth alpha channel $\alpha_{\text{gt}}$. The background mask is simply obtained by taking the inverse of the alpha mask:
\begin{equation}
M_{\text{bg}}(u, v) = 
    \begin{cases} 
        1, & \alpha_{\text{gt}}(u, v) = 0 \\ 
        0, & \text {otherwise} 
    \end{cases}
\end{equation}

\noindent\textbf{Anomaly Mask} ($M_{\text{anomaly}}$). We also construct a disparity anomaly mask to filter out non-positive or invalid values (such as Infinity or NaN) generated by the stereo model:
\begin{equation}
M_{\text{anomaly}}(u, v) = 
    \begin{cases} 
        1, & \hat{D}_{l}(u, v) \leq 0 \lor \hat{D}_{l}(u, v) = \text{Inf} \lor \hat{D}_{l}(u, v) = \text{NaN} \\ 
        0, & \text{otherwise} 
    \end{cases}
\end{equation}

\begin{figure}[htbp]
    \centering
    \includegraphics[width=\textwidth]{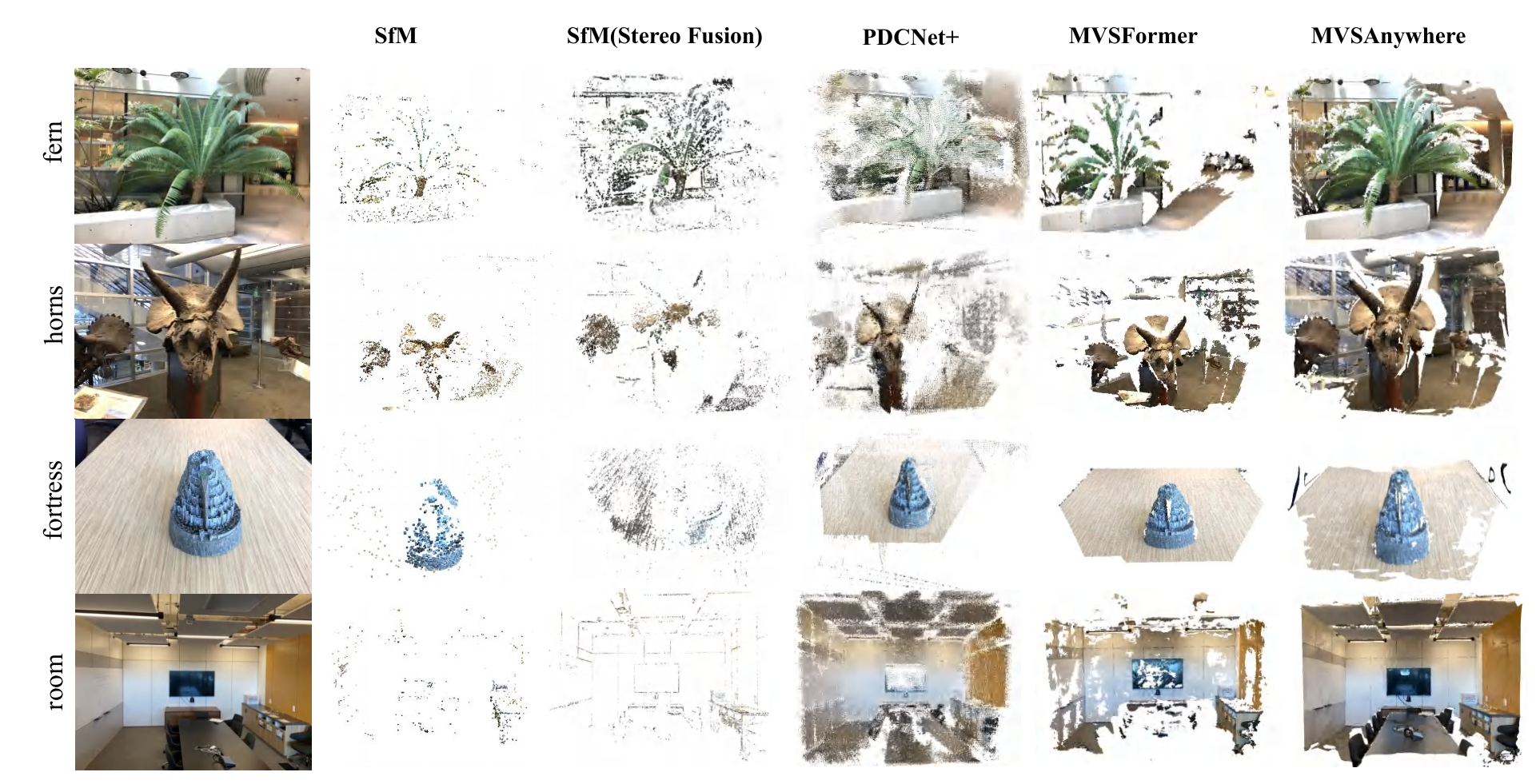}
    \caption{\textbf{Visual comparison of different Gaussian initialization methods.} We visualize the initial point clouds extracted from the \textit{fern}, \textit{horns}, \textit{fortress}, and \textit{room} scenes. Compared to the sparse SfM baseline and other advanced matching methods, MVSAnywhere provides significantly denser and structurally more complete point clouds, offering an optimal geometric prior for Gaussian optimization.}
    \label{fig:pcd_vis}
\end{figure}

\section{Additional Visualization Results}

\subsection{Visualization of Different Gaussian Initialization}

In this section, we provide comprehensive visual comparisons to demonstrate the effectiveness of our adopted initialization method, MVSAnywhere~\cite{izquierdo2025mvsanywhere}. As illustrated in Fig.~\ref{fig:pcd_vis}, we compare the initial point clouds generated by our method against traditional baselines and recent dense matching frameworks, including standard Structure-from-Motion~\cite{schonberger2016structure} (SfM), SfM with Stereo Fusion, PDCNet+~\cite{truong2023pdc}, and MVSFormer~\cite{cao2022mvsformer}.

The visualizations are evaluated across four distinct scenes exhibiting varying levels of geometric complexity: \textit{fern}, \textit{horns}, \textit{fortress}, and \textit{room}. As shown in the comparisons, traditional SfM often yields highly sparse representations, particularly in textureless regions. While other learning-based methods like PDCNet+ and MVSFormer improve the density, they may still struggle to recover complete fine-grained structures. In contrast, MVSAnywhere consistently generates the most complete, dense, and uniformly distributed point clouds across all scenes. This superior geometric initialization provides a highly robust prior for 3D Gaussian Splatting, which effectively minimizes floaters, accelerates the optimization process, and ultimately leads to higher-fidelity novel view synthesis.

\section{Trade-off between Training Overhead and Deployment Efficiency.}

While StereoGS incurs higher training costs due to its reliance on an external pre-trained model for online Stereo Depth Regularization, our Gradient-based Opacity Decay strategy mitigates this by substantially pruning the number of 3DGS primitives, thereby ensuring high deployment efficiency.

\begin{table}[htbp]
  \caption{\textbf{Average Training Overhead and 3DGS Count.}}
  \label{tab:training_overhead} 
  \centering
  \scriptsize
  \resizebox{\linewidth}{!}{ % 自动缩放至当前栏宽
      \begin{tabular}{l c c c c c c} 
        \toprule
        \multirow{2}{*}[-0.5ex]{Methods}
        & \multicolumn{2}{c}{Train Time (mins)}
        & \multicolumn{2}{c}{Peak VRAM (GB)} 
        & \multicolumn{2}{c}{3DGS Count} \\ 
        \cmidrule(lr){2-3} \cmidrule(lr){4-5} \cmidrule(l){6-7}

        & \multicolumn{1}{c}{LLFF (3-view)}
        & \multicolumn{1}{c}{DTU (3-view)}
        & \multicolumn{1}{c}{LLFF (3-view)}
        & \multicolumn{1}{c}{DTU (3-view)} 
        & \multicolumn{1}{c}{LLFF (3-view)}
        & \multicolumn{1}{c}{DTU (3-view)} \\ 
    
        \midrule
        3DGS & 4.3 & 3.9 & 0.5 & 0.3 & 265.1K & 168.6K \\
        FSGS & 26.5 & 15.6 & 11.8 & 3.7 & 370K & 71.6K \\
        MVPGS & 6.4 & 5.8 & 5.2 & 4.8 & 1521.7K & 220.9K \\
        \textbf{StereoGS (Ours)} & 43.2 & 34.6 & \textbf{3.6} & \textbf{2.5} & \textbf{126K} & \textbf{44.4K} \\
        \bottomrule
        \end{tabular}
    }
\end{table}

\end{document}